\documentclass[sigconf, screen]{acmart}
\AtBeginDocument{%
  }

\setcopyright{acmlicensed}
\copyrightyear{2026}
\acmYear{2026}
\acmDOI{XXXXXXX.XXXXXXX}
\acmConference[MM '26]{The 34th ACM International Conference on Multimedia}{November 10--14,
  2026}{Rio de Janeiro, Brazil}
\acmISBN{978-1-4503-XXXX-X/2018/06}

\acmSubmissionID{6388}
\renewcommand\footnotetextcopyrightpermission[1]{}
\usepackage{amsmath}
\usepackage{mathtools}
\usepackage{subcaption}
\usepackage[table]{xcolor}
\usepackage[capitalize,noabbrev]{cleveref}
\usepackage{multirow}
\usepackage{makecell}
\usepackage{siunitx}
\usepackage{algorithm}
\usepackage{algorithmic}

\providecommand{\triangleq}{\mathrel{\overset{\triangle}{=}}}

\begin{document}
\title{EvoESAP: Non-Uniform Expert Pruning for Sparse MoE}

\author{Zongfang Liu}
\authornote{Both authors contributed equally to this research.}
\affiliation{%
  \institution{Zhejiang University}
  \city{Hangzhou}
  \state{Zhejiang}
  \country{China}}
\affiliation{%
  \institution{Westlake University}
  \city{Hangzhou}
  \state{Zhejiang}
  \country{China}}

\author{Shengkun Tang}
\authornotemark[1]
\affiliation{%
  \institution{Mohamed bin Zayed University of Artificial Intelligence}
  \city{Abu Dhabi}
  \country{United Arab Emirates}}

\author{Boyang Sun}
\affiliation{%
  \institution{Mohamed bin Zayed University of Artificial Intelligence}
  \city{Abu Dhabi}
  \country{United Arab Emirates}}

\author{Ping Wang}
\affiliation{%
  \institution{Westlake University}
  \city{Hangzhou}
  \state{Zhejiang}
  \country{China}}

\author{Zhiqiang Shen}
\authornote{Corresponding authors.}
\affiliation{%
  \institution{Mohamed bin Zayed University of Artificial Intelligence}
  \city{Abu Dhabi}
  \country{United Arab Emirates}}
\email{zhiqiang.shen@mbzuai.ac.ae}

\author{Xin Yuan}
\authornotemark[2]
\affiliation{%
  \institution{Westlake University}
  \city{Hangzhou}
  \state{Zhejiang}
  \country{China}}
\email{xyuan@westlake.edu.cn}

\renewcommand{\shortauthors}{Anonymous Author(s)}

\newcommand{\RETURN}{\STATE \textbf{return}}
\sisetup{reset-text-series=false,text-series-to-math=true}

\begin{abstract}
Sparse Mixture-of-Experts (SMoE) language models achieve strong capability at low per-token compute, yet deployment remains constrained by memory footprint and throughput because the full expert pool must still be stored and served. Post-training expert pruning reduces this cost, but most methods focus on which experts to prune within each layer and default to a uniform layer-wise sparsity allocation, even though the layer-wise allocation can strongly affect performance. We decouple pruning into within-layer expert ranking and across-layer budget allocation, and introduce \textbf{E}xpected \textbf{S}peculative \textbf{A}cceptance \textbf{P}roxy (\textbf{ESAP}), a speculative-decoding-inspired, teacher-forced metric that measures how well a pruned model matches the full model without costly autoregressive decoding. ESAP is bounded and stable, enabling cheap comparison of many candidates. Building on ESAP, we propose EvoESAP, an evolutionary search framework that finds an improved non-uniform layer-wise sparsity allocation under a fixed global budget while holding the within-layer pruning order fixed, making it a plug-and-play method for criteria such as Frequency, EAN, SEER, and REAP. Across 7B--30B SMoE LLMs at 25\% and 50\% sparsity, EvoESAP consistently discovers non-uniform allocations that improve open-ended generation (up to \textbf{+19.6\%} on MATH-500 at 50\% sparsity) while preserving competitive multiple-choice accuracy compared with uniform pruning at the same sparsity. Code is available at \url{https://github.com/ZongfangLiu/EvoESAP}.
\end{abstract}



\maketitle

\section{Introduction}
\begin{figure*}
    \centering
    \includegraphics[width=.9\linewidth]{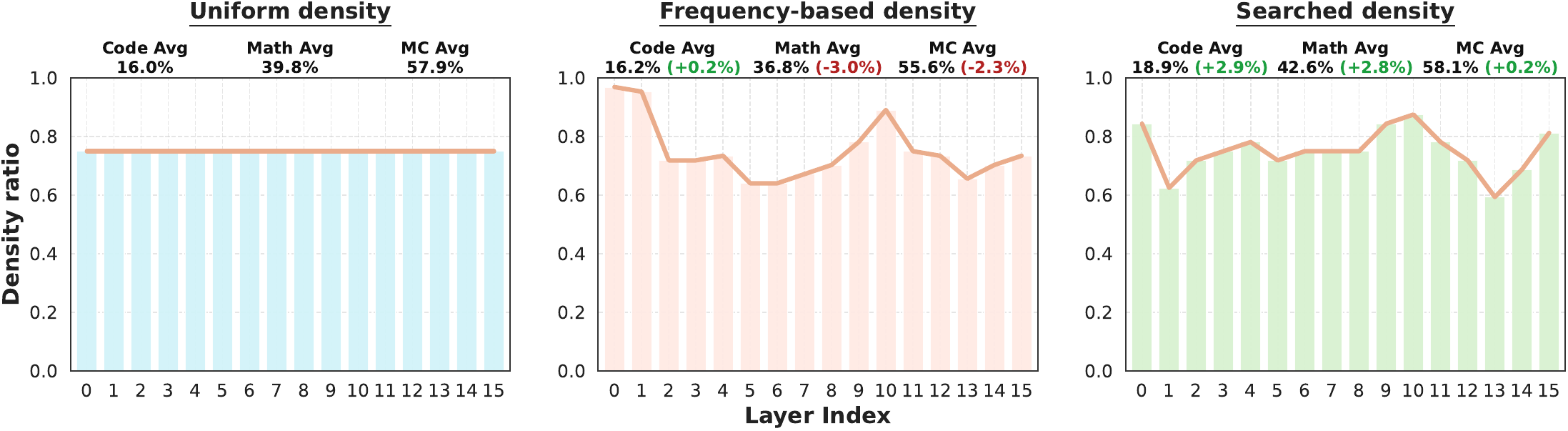}
    \caption{\textbf{Layer-wise density schedules and performance for OLMoE-1B-7B-0125-Instruct at 25\% global sparsity (density = $1-\text{sparsity}$).}
    Each panel shows the per-layer remaining expert density under a fixed global pruning budget, using REAP to rank experts in each layer (computed from 1,024 calibration samples from \texttt{evol-codealpaca-v1}).
    \textbf{Uniform} prunes the same fraction of experts in every layer.
    \textbf{Frequency-based} ranks experts globally by routing frequency, counts how many experts from each layer fall in the tail 25\%, and uses those counts to set layer-wise sparsity.
    \textbf{Searched} finds a non-uniform sparsity schedule with EvoESAP under the same budget. Numbers above panels report average performance on Code/Math/MC benchmarks, with deltas relative to uniform.
    The results imply that given the same pruning metric, non-uniform allocation has the potential to better preserve the model’s capabilities for SMoE expert pruning; however, finding an effective non-uniform allocation is non-trivial, and a poor allocation can harm overall performance.}
    \Description{Three panels compare uniform, frequency-based, and EvoESAP-searched layer-wise expert density schedules for OLMoE at 25 percent global sparsity, together with average coding, math, and multiple-choice results.}
    \label{fig:motivation_figure}
\end{figure*}

In Transformer architectures~\cite{vaswani2017attention}, Sparse Mixture-of-Experts (SMoE) models replace the dense feed-forward block with multiple expert FFNs and a learned router that dispatches each token to only its top-$k$ experts~\cite{shazeer2017outrageously}. This conditional computation enables overall parameter scaling while retaining low computation cost for each token. Recent SMoE LLMs deliver strong performance at lower per-token compute~\cite{jiang2024mixtral,muennighoff2024olmoe,liu2024deepseek,yang2025qwen3,meta2025llama,zeng2025glm,ernie2025technicalreport,team2025kimi}, but deployment remains costly because the full expert pool must be stored, stressing memory footprint and serving throughput. Routing analyses~\cite{huang2024mixture} further reveal expert-level redundancy and imbalanced expert usage at inference, suggesting room for expert-level compression~\cite{li2023merge,lu2024not,zhang2025diversifying,lee2025stun}.

Expert-level SMoE compression mainly follows two paradigms: expert merging and expert pruning. Recent evidence, however, points to an important limitation of merging: although it can preserve strong multiple-choice (MC) question-answering performance, it may harm open-ended generation quality, plausibly because merging introduces irreducible approximation error~\cite{lasby2025reap}. We therefore concentrate on expert pruning in this work. Notably, most existing studies assess compression primarily on MC benchmarks. Yet open-ended generation remains comparatively underexplored, despite more directly reflecting reasoning ability and text generation quality. Moreover, expert pruning entails two coupled decisions: determining which experts to retain within each layer and deciding how to distribute the pruning budget across layers. Most prior work emphasizes expert selection while defaulting to a uniform pruning ratio across layers. Yet evidence from both vision model pruning~\cite{lee2020layer,liu2022unreasonable} and dense LLM pruning~\cite{yin2023outlier,lu2024alphapruning,tang2025darwinlm} suggests that non-uniform layer-wise allocation can be crucial. For SMoE models, however, existing findings are inconclusive. SEER-MoE~\cite{muzio2024seer} reports that a non-uniform allocation derived from global frequency-based scores improves Mixtral-8$\times$7B on MC benchmarks, whereas MoE-$I^2$~\cite{yang2024moe} finds that, under a leave-one-out loss criterion, the optimal sparsity distribution for Mixtral is nearly uniform.

In this work, we aim to identify the pruned model whose generative behavior best matches that of the full model and to explore effective non-uniform layer-wise sparsity allocation. To assess generative similarity, a natural criterion is the speculative decoding acceptance rate: one can pair each pruned candidate with the full model and prefer the candidate with the highest acceptance. However, directly using speculative decoding for evaluation is computationally expensive because it requires autoregressive decoding for every candidate, which becomes increasingly inefficient as the candidate pool grows, as shown in Table~\ref{tab:specdec_vs_espdataset_code_search}. We therefore introduce Expected Speculative Acceptance Proxy (ESAP), a speculative-decoding-inspired, teacher-forced metric that measures how well a pruned model matches the full model. ESAP avoids costly autoregressive decoding and instead relies on teacher-forced likelihood evaluations, yielding a bounded, stable, and computationally efficient metric for comparing candidates. Furthermore, we show that, even under the same pruning metric and global sparsity budget, layer-wise allocation can strongly affect performance. As shown in \cref{fig:motivation_figure}, a well-chosen non-uniform schedule can improve performance, whereas a plausible heuristic can instead degrade it, making effective non-uniform allocation both important and non-trivial. 

To address this challenge, we propose \textbf{EvoESAP}, an evolutionary search framework for finding improved non-uniform layer-wise sparsity allocations under a fixed global budget. We decouple pruning into within-layer expert selection and across-layer budget allocation: given any expert-importance criterion, we first fix a per-layer pruning order, and then apply evolutionary search over allocations, using ESAP as the fitness function. As a plug-and-play method, EvoESAP can be applied to other heuristic pruning metrics such as Frequency, SEER~\citep{muzio2024seer}, EAN~\citep{jaiswal2025finding}, and REAP~\citep{lasby2025reap}. Empirically, we evaluate EvoESAP on OLMoE-1B-7B-0125-Instruct~\cite{muennighoff2024olmoe}, DeepSeek-V2-Lite-Chat~\cite{deepseekv2}, ERNIE-4.5-21B-A3B-PT~\cite{ernie2025technicalreport}, and Qwen3-30B-A3B-Instruct-2507~\cite{yang2025qwen3} at 25\% and 50\% global sparsity. Across four expert pruning metrics, EvoESAP consistently finds allocations that improve capability over uniform pruning under the same sparsity budget on generative benchmarks. Notably, at 50\% global sparsity on ERNIE-4.5-21B-A3B-PT, searching the allocation yields a \textbf{+19.6\%} gain on MATH-500 (vs.\ uniform sparsity allocation).

\noindent Our contributions can be summarized as:
\begin{itemize}\setlength\itemsep{2pt}
  \item We introduce ESAP, a speculative-decoding-inspired, teacher-forced proxy fitness function that enables efficient evaluation of pruning candidates for behavior preservation.
  \item We identify layer-wise budget allocation as a coupled yet under-studied decision in SMoE expert pruning: non-uniform schedules can help, whereas naive ones can hurt. Based on this finding, we propose EvoESAP, a plug-and-play evolutionary search framework that finds better non-uniform sparsity distributions under a fixed global budget while holding within-layer pruning orders fixed.
  \item Empirically, across four SMoE models at 25\% and 50\% sparsity and four pruning metrics, EvoESAP consistently improves over uniform allocation under the same global budget, with the largest gain reaching \textbf{+19.6\%} on MATH-500 for ERNIE-4.5-21B-A3B-PT at 50\% global sparsity.

\end{itemize}

\section{Related Work}
\subsection{Expert Pruning}
Early work on SMoE pruning is often task- or domain-specific, producing smaller specialized models by removing experts that are rarely useful in a target setting~\cite{chen2022task,koishekenov2023memory}. More recent studies instead consider task-agnostic post-training pruning for MoE LLMs. NAEE~\cite{lu2024not} highlights substantial expert imbalance and proposes expert dropping/skipping policies guided by router or expert signals. EEP~\cite{liu2024efficient} shows that evolutionary search can be used to optimize which experts to prune under a fixed uniform per-layer pruning ratio, and then employs expert merging to recover knowledge from the pruned experts. STUN~\cite{lee2025stun} proposes a structured-then-unstructured MoE pruning pipeline that clusters experts to prune redundant ones 
and then applies unstructured pruning \cite{sun2023simple, yin2023outlier} within the surviving experts. 
Most SMoE pruning works are evaluated mainly on multiple-choice benchmarks, while open-ended generation is largely neglected. REAP~\cite{lasby2025reap} argues that pruning better preserves open-ended generation than merging, and their router-weighted expert activation score outperforms frequency-based pruning and the Expert Activation Norm (EAN); notably, EAN is the best among 16 criteria evaluated in~\cite{jaiswal2025finding}. However, most existing methods implicitly assume uniform sparsity across layers. To the best of our knowledge, explicit discussion of non-uniform sparsity allocation for SMoE pruning remains limited: SEER-MoE~\cite{muzio2024seer} empirically suggests that, when allocating sparsity using frequency-based scores, global non-uniform sparsity allocation can outperform uniform on MMLU~\cite{hendrycks2020measuring} for Mixtral-8$\times$7B~\cite{jiang2024mixtral}. DiEP~\cite{bai2025diepadaptivemixtureofexpertscompression} proposes a differentiable optimization framework that jointly models expert pruning and non-uniform layer-wise sparsity allocation. However, because within-layer expert selection and across-layer allocation are optimized in an entangled manner, the contribution of non-uniform allocation itself is hard to disentangle. Moreover, its evaluation is limited to four MC benchmarks, leaving open-ended generation unexamined. Thus, whether and how non-uniform sparsity allocation benefits SMoE pruning---especially for open-ended generation---remains an open question.

\subsection{Expert Merging}
Early work on expert merging is exemplified by MEO~\cite{he2023merging}, which dynamically merges the activated experts into a single expert at inference time, using router (gating) scores as the weights. In contrast, more recent merging methods are typically static and rely on clustering to identify redundant experts to consolidate. MC-SMoE~\cite{li2023merge} leverages routing statistics to decide which experts to merge: it first performs neuron permutation alignment, then identifies globally dominant experts and assigns the remaining experts as “group members” based on routing behavior, and finally merges each group via activation-frequency-weighted averaging. HC-SMoE~\cite{chen2024retraining} clusters experts by their output similarity on a calibration set and applies hierarchical clustering to form robust groups; experts within each cluster are then merged using frequency-weighted averaging. DERN~\cite{zhou2025dropping} extends the granularity from the expert level to the segment level: it first drops redundant experts based on router statistics, then decomposes the dropped experts into neuron-level segments and reassigns these segments to compatible retained experts for merging. As reported in~\cite{lasby2025reap}, while the state-of-the-art merging method HC-SMoE generally outperforms fine-tuning-free expert pruning methods on MC benchmarks, it can suffer a substantial performance drop on open-ended generation tasks.

\subsection{Other Compression Methods}
Beyond whole-expert pruning and merging, SMoE models can also be compressed at a finer granularity through quantization~\cite{huang2024mixture}, decomposition-based compression~\cite{gu2025delta,he2025efficiently,li2025moe}, intra-expert weight pruning as in MoE-Pruner~\cite{xie2024moe}, and subspace-level expert merging as in Sub-MoE~\cite{li2025sub}. A growing line of work further combines multiple techniques~\cite{he2024towards,liu2024survey}. For example, MoE-I$^2$~\cite{yang2024moe} integrates expert pruning and low-rank decomposition, followed by LoRA fine-tuning~\cite{hu2022lora} to recover performance. In its pruning stage, MoE-I$^2$ estimates layer importance using a leave-one-out loss increase from removing each expert, allocates per-layer pruning budgets accordingly, and then applies a genetic search to select the experts to retain per layer. Under this criterion, it reports nearly uniform layer-wise sparsity for Mixtral-8$\times$7B~\cite{jiang2024mixtral} and Qwen1.5-MoE-A2.7B~\cite{qwen_moe}, but a highly non-uniform pattern for DeepSeek-V2-Lite~\cite{deepseekv2}, contrasting with SEER-MoE where non-uniform allocation benefits Mixtral-8$\times$7B. Taken together, prior findings paint an inconsistent picture of whether layer-wise sparsity should be uniform or non-uniform across SMoE, leaving the question unresolved.

\section{Method}

\subsection{Sparse Mixture-of-Experts Architecture}
Sparse Mixture-of-Experts (SMoE) layers introduce conditional computation by replacing a dense feed-forward network (FFN)
with a set of $n$ expert FFNs $\{E_i\}_{i=1}^{n}$ and a router (gating) network that selects only a few experts per token~\cite{shazeer2017outrageously}.
Given a token hidden state $h\in\mathbb{R}^d$, the router produces logits
$z = \mathbf{W}_g h \in \mathbb{R}^{n}$, where $\mathbf{W}_g$ is the router projection.
Let $\mathcal{A}(h)=\mathrm{TopK}(z,k)$ denote the indices of the $k$ largest logits ($k\ll n$). The sparse gating weights are
\begin{equation}
g_i(h)=
\begin{cases}
\dfrac{\exp(z_i)}{\sum_{j\in \mathcal{A}(h)}\exp(z_j)}, & i\in \mathcal{A}(h),\\[6pt]
0, & i\notin \mathcal{A}(h),
\end{cases}
\label{eq:topk_gate}
\end{equation}
and the MoE output is the weighted mixture of the selected experts:
\begin{equation}
y(h) \;=\; \sum_{i\in \mathcal{A}(h)} g_i(h)\,E_i(h).
\label{eq:moe_output}
\end{equation}
Because $g(h)$ has only $k$ nonzero entries, only the selected experts are activated, so the inference latency is approximately proportional to $k$ while total capacity scales with the number of experts~\cite{fedus2022switch}.

\subsection{Problem Formulation}
\label{sec:budget_prune}

The goal of expert pruning is to obtain a compressed SMoE model that preserves the full model’s behavior while reducing deployment cost (e.g., memory usage). This entails two coupled choices: which experts to remove (within-layer selection) and how many to remove per layer (across-layer allocation). Most prior work focuses on selection while implicitly adopting uniform per-layer pruning, thereby overlooking the allocation axis. We therefore decouple pruning into two steps: we first fix a per-layer pruning order (e.g., from REAP), and then use an evolutionary search to optimize a non-uniform layer-wise allocation under the same global budget, guided by ESAP (\cref{sec:esap}), the proposed speculative-decoding-inspired, behavior-preserving fitness function.

\noindent \textbf{Per-layer pruning order.}
Consider an SMoE model with $L$ MoE layers. Layer $\ell \in \{1,\dots,L\}$ contains $n_\ell$ experts
$\{E_{\ell,1},\dots,E_{\ell,n_\ell}\}$ (typically $n_\ell \equiv n$ in practice).
We compute an importance score $s_{\ell,j}$ for each expert $E_{\ell,j}$ using a chosen criterion (e.g., REAP), and define a
layer-wise pruning order by sorting experts in ascending importance:
$\pi_\ell \;=\; \mathrm{argsort}_{j \in \{1,\dots,n_\ell\}} \big(s_{\ell,j}\big)$
where $\pi_\ell$ is a permutation of $\{1,\dots,n_\ell\}$ such that
$s_{\ell,\pi_\ell(1)} \le \cdots \le s_{\ell,\pi_\ell(n_\ell)}$. Given a layer-wise pruning budget $r_\ell \in \{0,1,\dots,n_\ell-k_\ell\}$ (to ensure at least $k_\ell$ experts remain for top-$k_\ell$
routing), we prune the $r_\ell$ least important experts:
\begin{equation}
\mathcal{P}_\ell(r_\ell) \;=\; \{\pi_\ell(j)\}_{j=1}^{r_\ell}.
\label{eq:pruned_set}
\end{equation}

\subsection{Level-Switch Evolutionary Search}
\label{sec:evo_search}
\noindent \textbf{Search space, constraint, and objective.}
With the per-layer pruning orders $\{\pi_\ell\}$ fixed, the remaining degree of freedom is the layer-wise sparsity allocation
under a fixed global pruning budget. We therefore search over integer allocation vectors $\mathbf{r}=(r_1,\dots,r_L)$, where
$r_\ell$ is the number of experts removed in layer $\ell$ and $B$ is the \emph{global} pruning budget (total number of experts removed across all MoE layers).
Concretely, we solve
\begin{equation}
\begin{gathered}
\mathbf{r}^\star \in \arg\max_{\mathbf{r}\in\mathbb{Z}^L} f(\mathbf{r};\mathcal{D}_{\text{search}}) \\
\text{s.t.}\ \sum_{\ell=1}^{L} r_\ell = B,\quad 0 \le r_\ell \le n_\ell - k_\ell,\ \forall \ell,
\end{gathered}
\label{eq:allocation_objective}
\end{equation}
so that at least $k_\ell$ experts remain in every layer for top-$k_\ell$ routing.
The search set $\mathcal{D}_{\text{search}}$ is used only for ESAP fitness evaluation; the pruning orders $\{\pi_\ell\}$ are computed once
from the chosen importance criterion on its own calibration data.

\noindent \textbf{Initialization.}
We initialize population $\mathcal{S}^{(0)}$ of size $P$ with a mixture of structured patterns and random feasible allocations:
(i) a uniform allocation, (ii) several patterned allocations that concentrate pruning in different layer regions (e.g., early-heavy,
middle-heavy, late-heavy), and (iii) the remaining individuals sampled uniformly from all feasible allocations that satisfy the global budget.

\noindent \textbf{Selection.}
We evaluate each candidate $\mathbf{r}\in\mathcal{S}^{(t)}$ and keep the top $m$ survivors:
\begin{equation}
\mathcal{S}^{(t)}_{\mathrm{elite}}
\;=\;
\mathrm{TopM}\big(\{\mathbf{r} \in \mathcal{S}^{(t)}\};\, f(\mathbf{r}),\, m\big),
\end{equation}
where $\mathrm{TopM}$ returns the $m$ highest-fitness candidates. We carry $\mathcal{S}^{(t)}_{\mathrm{elite}}$ into the next
generation and apply mutation on them to generate $P-m$ offspring, thereby keeping the population size constant.

\noindent \textbf{Level-switch mutation.}
To generate an offspring, we sample a parent $\mathbf{r}$ from $\mathcal{S}^{(t)}_{\mathrm{elite}}$ and apply a budget-preserving
\emph{level-switch} operator that transfers pruning budget between two layers while keeping the global constraint unchanged.
A single level-switch step samples two distinct layers $a\neq b$ and a transfer size $\Delta\in\{1,\dots,\Delta_{\max}\}$, and updates
\begin{equation}
r_\ell' \;=\;
\begin{cases}
r_\ell + \Delta, & \ell=a,\\
r_\ell - \Delta, & \ell=b,\\
r_\ell, & \text{otherwise},
\end{cases}
\label{eq:level_switch}
\end{equation}
subject to feasibility $0 \le r_b - \Delta$ and $r_a + \Delta \le n_a - k_a$.
If the sampled $(a,b,\Delta)$ is infeasible, we resample until \cref{eq:level_switch} is valid.

We apply this operator multiple times per offspring by composing $\tau$ feasible level-switch steps, where the mutation count is
\begin{equation}
\tau \;=\; \min\!\big(U\{1,\dots,\tau_{\max}\},\,U\{1,\dots,\tau_{\max}\}\big),
\end{equation}
$U\{1,\dots,\tau_{\max}\}$ denotes a discrete uniform draw over that set, and the two draws are independent. This choice biases mutations toward small local reallocations (exploitation), while still occasionally producing larger jumps through multiple transfers (exploration).

\noindent \textbf{Termination and output.}
We run the search for $T$ generations and output the best-found allocation
\begin{equation}
\mathbf{r}^\star \;=\; \arg\max_{\mathbf{r}\in \cup_{t=0}^{T}\mathcal{S}^{(t)}} f(\mathbf{r};\mathcal{D}_{\text{search}}),
\end{equation}
and instantiate the pruned model by applying $\{\mathcal{P}_\ell(r_\ell^\star)\}_{\ell=1}^{L}$ from \cref{eq:pruned_set}. Pseudocode of searching is provided in \cref{app:evo_alg}.

\subsection{Expected Speculative Acceptance Proxy}
\label{sec:esap}

\begin{figure}[t]
  \centering
  \begin{subfigure}[t]{0.49\linewidth}
    \centering
    \includegraphics[width=0.87\linewidth]{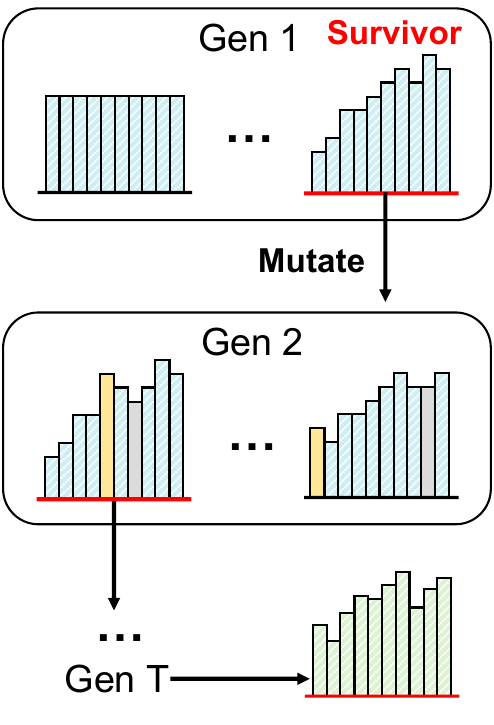}
    \caption{Search overview.}
  \end{subfigure}\hfill
  \begin{subfigure}[t]{0.49\linewidth}
    \centering
    \includegraphics[width=0.87\linewidth]{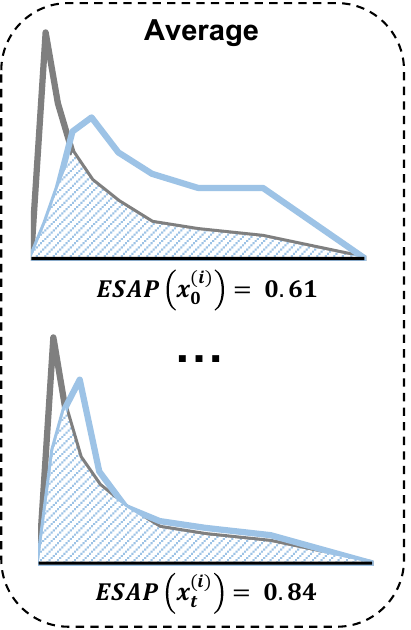}
    \caption{ESAP$^{(i)}$.}
  \end{subfigure}
\caption{\textbf{Overview of EvoESAP.}
\textbf{(a) Evolutionary search with budget-preserving level-switch mutation.}
Histograms visualize the layer-wise density distribution of each candidate model (density $=1-\,$sparsity) induced by an allocation $\mathbf{r}$
(experts removed per layer) under a fixed global budget $B$.
Offspring are generated from the top $m$ survivors by a \emph{level-switch} that transfers $\Delta$ units of pruning budget
between two layers (gray: decrease; yellow: increase), keeping $\sum_{\ell} r_{\ell}=B$ unchanged.
\textbf{(b) ESAP as per-sample fitness.}
Under teacher forcing, ESAP scores a sample by the full-vocabulary overlap between the target next-token
distribution (dark gray, $p(\cdot\mid x)$) and the candidate/draft distribution (blue, $q(\cdot\mid x)$),
averaged over answer-token positions. The ESAP score is defined in \cref{eq:esap_def}, and higher values are better.
}
\Description{The left panel shows evolutionary search over layer-wise sparsity allocations under a fixed pruning budget. The right panel illustrates ESAP as overlap between target and candidate next-token distributions under teacher forcing.}
\label{fig:search_and_esap}

\end{figure}

In our evolutionary search, we seek a non-uniform layer-wise sparsity allocation that keeps the pruned candidate closer to the full model. Accordingly, our fitness should (i) correlate with behavioral similarity to the full model and (ii) be cheap enough to evaluate for hundreds/thousands of candidates.

\noindent \textbf{Speculative decoding as a compatibility signal.}
Speculative decoding~\cite{leviathan2023fast, chen2023accelerating} accelerates inference by letting a lightweight draft model propose tokens, while the target model verifies and either accepts them or falls back to its own token. The more similar the outputs of the two models are, the faster the inference.
Intuitively, if a candidate can serve as an effective draft model for its full version, then its behavior should be close to that of the full model, suggesting the speculative acceptance rate as a natural compatibility signal.
Let $p(\cdot \mid x)$ denote the target next-token distribution given context (prefix) $x$,
and let $q(\cdot \mid x)$ denote the candidate/draft distribution.
When the draft proposes a token $y \sim q(\cdot \mid x)$, the standard acceptance probability is
\begin{equation}
\alpha(x,y) \;=\; \min\!\left(1,\frac{p(y\mid x)}{q(y\mid x)}\right),
\label{eq:spec_accept}
\end{equation}
together with the standard rejection correction step guarantees that the overall decoded token distribution exactly matches the target model \cite{leviathan2023fast}.
However, directly estimating speculative acceptance inside an inner-loop search is expensive:
it requires autoregressive generation for many prompts, running \emph{both} models, and the measured acceptance depends on the sampled trajectory, block size, and prefix drift after rejections. As shown in \cref{tab:specdec_vs_espdataset_code_search}, this is prohibitively expensive and introduces substantial variance, making it unsuitable for ranking hundreds or thousands of candidates.

\noindent \textbf{SAP: a dataset-level single-token acceptance proxy.}
A natural way to remove trajectory dependence while staying close to the speculative-decoding acceptance test is to evaluate
\cref{eq:spec_accept} on fixed, teacher-forced contexts from a search set.
Let $\mathcal{D}_{\text{search}}=\{(u^{(i)},a^{(i)})\}_{i=1}^{N}$ contain $N$ prompt--answer pairs. For each sample $i$, we teacher-force on $(u^{(i)},a^{(i)})$ but score only answer-token positions: $\mathcal{I}^{(i)}$ indexes the next-token contexts whose \emph{next token} lies in $a^{(i)}$. For each $t\in\mathcal{I}^{(i)}$, let $x_t^{(i)}$ be the corresponding prefix context, and let
$p(\cdot\mid x_t^{(i)})$ and $q(\cdot\mid x_t^{(i)})$ denote the target and candidate next-token distributions.

At a single context $x_t^{(i)}$, the Speculative Acceptance Proxy (SAP) draws one proposal
$\hat{y}_t^{(i)}\sim q(\cdot\mid x_t^{(i)})$ and evaluates its speculative acceptance probability:
\begin{equation}
\mathrm{SAP}\!\left(x_t^{(i)}\right)
\;\triangleq\;
\min\!\left(1,\frac{p(\hat{y}_t^{(i)}\mid x_t^{(i)})}{q(\hat{y}_t^{(i)}\mid x_t^{(i)})}\right).
\label{eq:sap_def}
\end{equation}
We average over answer positions within each sample,
\begin{equation}
\mathrm{SAP}^{(i)}
\;\triangleq\;
\frac{1}{|\mathcal{I}^{(i)}|}\sum_{t\in\mathcal{I}^{(i)}} \mathrm{SAP}\!\left(x_t^{(i)}\right),
\end{equation}
and then report the dataset-level score as the mean over samples:
\begin{equation}
\mathrm{SAP}
\;\triangleq\;
\frac{1}{N}\sum_{i=1}^{N} \mathrm{SAP}^{(i)}.
\label{eq:sap_dataset}
\end{equation}

Since SAP is evaluated on shared, fixed teacher-forced contexts (answer tokens only), it avoids prefix drift and reduces computation time. However, it remains a Monte-Carlo estimator since it depends on the sampled proposal
$\hat{y}_t^{(i)}$, which can introduce variance.

\begin{table*}[tbp]
  \centering
  \caption{Benchmark results for OLMoE-1B-7B-0125-Instruct and ERNIE-4.5-21B-A3B-PT on coding, WildBench, math, and MC benchmarks. OLMoE is evaluated at 25\% global sparsity, and ERNIE at 25\% and 50\%. \textsc{Uni} uses uniform layer-wise sparsity, while \textsc{ESAP} uses the non-uniform allocation found by EvoESAP. Bold indicates the better result within each pair.}
  \label{tab:main_table_olmoe_ernie}
  \resizebox{0.84\textwidth}{!}{%
  \begin{tabular}{c l l l | c c c | c | c c c | c}
    \toprule
    \multicolumn{4}{c}{} & \multicolumn{3}{c|}{\textbf{Coding}} & \multicolumn{1}{c|}{\textbf{Creative Writing}} & \multicolumn{3}{c|}{\textbf{Math}} & \multicolumn{1}{c}{\textbf{MC}} \\
    \textbf{Model} & \textbf{Sparsity} & \textbf{Method} &  &
    \textbf{Eval+} & \textbf{LiveCode} & \textbf{Avg} &
    \textbf{WildBench} &
    \textbf{GSM8K} & \textbf{MATH-500} & \textbf{Avg} &
    \textbf{MC Avg} \\
    \midrule

    \multirow[c]{10}{*}{OLMoE}
      & \multicolumn{3}{l|}{\textbf{Full}} & 0.341 & 0.033 & 0.187 & 0.444 & 0.682 & 0.222 & 0.452 & 0.653 \\
    \cmidrule(lr){2-12}
      & \multirow{8}{*}{25\%}
      & \multirow{2}{*}{EAN} & \textsc{Uni}   & \textbf{0.343} & 0.022 & \textbf{0.183} & \textbf{0.269} & \textbf{0.585} & 0.190 & 0.387 & \textbf{0.551} \\
      &                       &                   & \textsc{ESAP}  & 0.312 & \textbf{0.027} & 0.170 & 0.258 & 0.576 & \textbf{0.232} & \textbf{0.404} & 0.543    \\
      \cmidrule(lr){3-12}
      &                       & \multirow{2}{*}{SEER} & \textsc{Uni}  & \textbf{0.339} & \textbf{0.027} & \textbf{0.183} & 0.253 & 0.577 & 0.204 & 0.390 & \textbf{0.545} \\
      &                       &                       & \textsc{ESAP} & 0.306 & 0.022 & 0.164 & \textbf{0.254} & \textbf{0.601} & \textbf{0.248} & \textbf{0.424} & 0.539    \\
      \cmidrule(lr){3-12}
      &                       & \multirow{2}{*}{Freq} & \textsc{Uni}  & 0.341 & \textbf{0.022} & \textbf{0.182} & \textbf{0.265} & 0.591 & \textbf{0.220} & \textbf{0.405} & \textbf{0.547} \\
      &                       &                             & \textsc{ESAP} & \textbf{0.342} & \textbf{0.022} & \textbf{0.182} & 0.244 & \textbf{0.596} & 0.208 & 0.402 & 0.539 \\
      \cmidrule(lr){3-12}
      &                       & \multirow{2}{*}{REAP} & \textsc{Uni}   & 0.314 & 0.005 & 0.160 & \textbf{0.292} & 0.596 & 0.200 & 0.398 & 0.579 \\
      &                       &                        & \textsc{ESAP}  & \textbf{0.344} & \textbf{0.033} & \textbf{0.189} & 0.279 & \textbf{0.636} & \textbf{0.216} & \textbf{0.426} & \textbf{0.581} \\

    \midrule[\heavyrulewidth]

    \multirow[c]{21}{*}{ERNIE}
      & \multicolumn{3}{l|}{\textbf{Full}} & 0.867 & 0.247 & 0.557 & 0.479 & 0.829 & 0.780 & 0.804 & 0.721 \\
    \cmidrule(lr){2-12}
      & \multirow{8}{*}{25\%}
      & \multirow{2}{*}{EAN} & \textsc{Uni}   & 0.827 & 0.214 & 0.520 & \textbf{0.377} & 0.815 & 0.748 & 0.781 & 0.669 \\
      &                       &                   & \textsc{ESAP}  & \textbf{0.832} & \textbf{0.225} & \textbf{0.528} & 0.333 & \textbf{0.823} & \textbf{0.772} & \textbf{0.797} & \textbf{0.675} \\
      \cmidrule(lr){3-12}
      &                       & \multirow{2}{*}{\makecell[l]{SEER}} & \textsc{Uni}  & 0.830 & \textbf{0.214} & \textbf{0.522} & \textbf{0.301} & \textbf{0.804} & \textbf{0.736} & \textbf{0.770} & 0.634 \\
      &                       &                       & \textsc{ESAP} & \textbf{0.838} & 0.203 & 0.520 & 0.291 & \textbf{0.804} & 0.728 & 0.766 & \textbf{0.638} \\
      \cmidrule(lr){3-12}
      &                       & \multirow{2}{*}{Freq} & \textsc{Uni}  & \textbf{0.818} & 0.181 & 0.499 & 0.314 & 0.810 & 0.692 & 0.751 & 0.636 \\
      &                       &                             & \textsc{ESAP} & 0.811 & \textbf{0.236} & \textbf{0.524} & \textbf{0.316} & \textbf{0.815} & \textbf{0.716} & \textbf{0.765} & \textbf{0.647} \\
      \cmidrule(lr){3-12}
      &                       & \multirow{2}{*}{REAP} & \textsc{Uni}   & 0.823 & \textbf{0.231} & \textbf{0.527} & 0.354 & \textbf{0.821} & 0.730 & 0.775 & 0.667 \\
      &                       &                        & \textsc{ESAP}  & \textbf{0.833} & 0.209 & 0.521 & \textbf{0.376} & 0.814 & \textbf{0.752} & \textbf{0.783} & \textbf{0.672} \\
    \cmidrule(lr){2-12}
      & \multirow{8}{*}{50\%}
      & \multirow{2}{*}{EAN} & \textsc{Uni}   & 0.636 & \textbf{0.148} & 0.392 & 0.156 & \textbf{0.748} & 0.542 & 0.645 & \textbf{0.585} \\
      &                       &                   & \textsc{ESAP}  & \textbf{0.659} & 0.143 & \textbf{0.401} & \textbf{0.186} & 0.744 & \textbf{0.558} & \textbf{0.651} & 0.582 \\
      \cmidrule(lr){3-12}
      &                       & \multirow{2}{*}{\makecell[l]{SEER}} & \textsc{Uni}  & 0.698 & 0.170 & 0.434 & 0.130 & 0.555 & 0.418 & 0.487 & \textbf{0.551} \\
      &                       &                       & \textsc{ESAP} & \textbf{0.709} & \textbf{0.181} & \textbf{0.445} & \textbf{0.151} & \textbf{0.640} & \textbf{0.508} & \textbf{0.574} & 0.547 \\
      \cmidrule(lr){3-12}
      &                       & \multirow{2}{*}{Freq} & \textsc{Uni}  & \textbf{0.647} & \textbf{0.143} & \textbf{0.395} & 0.130 & 0.522 & 0.272 & 0.397 & \textbf{0.565} \\
      &                       &                             & \textsc{ESAP} & \textbf{0.647} & 0.126 & 0.387 & \textbf{0.151} & \textbf{0.631} & \textbf{0.468} & \textbf{0.549} & 0.557 \\
      \cmidrule(lr){3-12}
      &                       & \multirow{2}{*}{REAP} & \textsc{Uni}   & 0.730 & \textbf{0.192} & 0.461 & \textbf{0.215} & 0.695 & \textbf{0.598} & 0.646 & \textbf{0.575} \\
      &                       &                        & \textsc{ESAP}  & \textbf{0.737} & 0.187 & \textbf{0.462} & 0.205 & \textbf{0.718} & 0.578 & \textbf{0.648} & \textbf{0.575} \\

    \bottomrule
  \end{tabular}%
  }
\end{table*}

\noindent \textbf{ESAP: an expected acceptance proxy of speculative decoding.}
SAP in \cref{eq:sap_def} depends on a sampled proposal $\hat{y}_t^{(i)}\sim q(\cdot\mid x_t^{(i)})$, which can introduce
variance. We remove this Monte-Carlo noise by taking the expectation of the same acceptance test under the draft distribution.
For each teacher-forced context $x_t^{(i)}$, we define
\begin{equation}
\mathrm{ESAP}\!\left(x_t^{(i)}\right)
\;\triangleq\;
\mathbb{E}_{y \sim q(\cdot \mid x_t^{(i)})}
\left[\min\!\left(1,\frac{p(y\mid x_t^{(i)})}{q(y\mid x_t^{(i)})}\right)\right].
\label{eq:esap_def}
\end{equation}
Expanding the expectation yields a closed form:
\begin{align}
\mathrm{ESAP}\!\left(x_t^{(i)}\right)
&= \sum_{v \in \mathcal{V}} q(v\mid x_t^{(i)})\,\min\!\left(1,\frac{p(v\mid x_t^{(i)})}{q(v\mid x_t^{(i)})}\right)
\label{eq:esap_expand}\\
&= \sum_{v \in \mathcal{V}} \min\!\big(p(v\mid x_t^{(i)}),\,q(v\mid x_t^{(i)})\big),
\label{eq:esap_min}
\end{align}
using $q(v)\min(1,p(v)/q(v))=\min(p(v),q(v))$ for each vocabulary token $v$. Analogous to \cref{eq:sap_dataset}, we average ESAP over answer positions within each sample:
\begin{equation}
\mathrm{ESAP}^{(i)}
\;\triangleq\;
\frac{1}{|\mathcal{I}^{(i)}|}\sum_{t\in\mathcal{I}^{(i)}} \mathrm{ESAP}\!\left(x_t^{(i)}\right),
\end{equation}
and report the dataset-level score as
\begin{equation}
\mathrm{ESAP}
\;\triangleq\;
\frac{1}{N}\sum_{i=1}^{N} \mathrm{ESAP}^{(i)}.
\label{eq:esap_dataset}
\end{equation}
\cref{fig:search_and_esap}(b) visualizes $\mathrm{ESAP}^{(i)}$ as the token-level overlap between
$p(\cdot\mid x_t^{(i)})$ and $q(\cdot\mid x_t^{(i)})$ averaged over answer positions. Moreover, since $\mathrm{ESAP}(x)=\sum_{v\in\mathcal{V}} \min\!\big(p(v\mid x),\,q(v\mid x)\big),$
it also can be viewed as the complement of total variation:
\begin{equation}
\mathrm{ESAP}(x)
= 1 - \mathrm{TV}\!\left(p(\cdot\mid x),\,q(\cdot\mid x)\right),
\label{eq:esap_tv}
\end{equation}
where $\mathrm{TV}(p,q)\coloneqq \tfrac12\sum_{v\in\mathcal{V}}|p(v)-q(v)|$
(a detailed derivation is provided in \cref{app:esap_tv} for completeness).

\begin{table*}[tbp]
  \centering
  \caption{Benchmark results for DeepSeek-V2-Lite-Chat and Qwen3-30B-A3B-Instruct-2507 on coding, WildBench, math, and multiple-choice benchmarks. DeepSeek is evaluated at 25\% global sparsity, while Qwen3 is evaluated at 25\% and 50\%. \textsc{Uni} uses uniform layer-wise sparsity, while \textsc{ESAP} uses the non-uniform allocation found by EvoESAP. }
  \label{tab:main_table_deepseek_qwen3}
  \resizebox{0.84\textwidth}{!}{%
  \begin{tabular}{c l l l | c c c | c | c c c | c}
    \toprule
    \multicolumn{4}{c}{} & \multicolumn{3}{c|}{\textbf{Coding}} & \multicolumn{1}{c|}{\textbf{Creative Writing}} & \multicolumn{3}{c|}{\textbf{Math}} & \multicolumn{1}{c}{\textbf{MC}} \\
    \textbf{Model} & \textbf{Sparsity} & \textbf{Method} &  &
    \textbf{Eval+} & \textbf{LiveCode} & \textbf{Avg} &
    \textbf{WildBench} &
    \textbf{GSM8K} & \textbf{MATH-500} & \textbf{Avg} &
    \textbf{MC Avg} \\
    \midrule

    \multirow[c]{10}{*}{DeepSeek}
      & \multicolumn{3}{l|}{\textbf{Full}} & 0.549 & 0.104 & 0.327 & 0.418 & 0.610 & 0.298 & 0.454 & 0.678 \\
    \cmidrule(lr){2-12}
      & \multirow{8}{*}{25\%}
      & \multirow{2}{*}{EAN} & \textsc{Uni}   & \textbf{0.459} & \textbf{0.071} & \textbf{0.265} & 0.259 & 0.531 & \textbf{0.224} & 0.378 & 0.571 \\
      &                       &                        & \textsc{ESAP}  & 0.417 & \textbf{0.071} & 0.244 & \textbf{0.301} & \textbf{0.572} & \textbf{0.224} & \textbf{0.398} & \textbf{0.588} \\
      \cmidrule(lr){3-12}
      &                       & \multirow{2}{*}{SEER} & \textsc{Uni}   & 0.404 & 0.099 & 0.252 & 0.260 & 0.553 & 0.210 & 0.382 & 0.586 \\
      &                       &                        & \textsc{ESAP}  & \textbf{0.460} & \textbf{0.115} & \textbf{0.288} & \textbf{0.290} & \textbf{0.559} & \textbf{0.226} & \textbf{0.393} & \textbf{0.594} \\
      \cmidrule(lr){3-12}
      &                       & \multirow{2}{*}{Freq} & \textsc{Uni}   & \textbf{0.297} & \textbf{0.099} & \textbf{0.198} & 0.234 & 0.387 & 0.158 & 0.273 & 0.560 \\
      &                       &                        & \textsc{ESAP}  & 0.292 & 0.082 & 0.187 & \textbf{0.252} & \textbf{0.405} & \textbf{0.166} & \textbf{0.286} & \textbf{0.588} \\
      \cmidrule(lr){3-12}
      &                       & \multirow{2}{*}{REAP} & \textsc{Uni}   & \textbf{0.473} & \textbf{0.077} & \textbf{0.275} & 0.301 & 0.545 & 0.222 & 0.384 & 0.604 \\
      &                       &                        & \textsc{ESAP}  & 0.442 & 0.071 & 0.257 & \textbf{0.314} & \textbf{0.571} & \textbf{0.224} & \textbf{0.397} & \textbf{0.639} \\
    \midrule[\heavyrulewidth]

    \multirow[c]{21}{*}{Qwen3}
      & \multicolumn{3}{l|}{\textbf{Full}} & 0.871 & 0.368 & 0.619 & 0.644 & 0.923 & 0.802 & 0.863 & 0.737 \\
    \cmidrule(lr){2-12}
      & \multirow{8}{*}{25\%}
      & \multirow{2}{*}{EAN} & \textsc{Uni}   & \textbf{0.871} & \textbf{0.363} & \textbf{0.617} & \textbf{0.517} & \textbf{0.902} & 0.748 & 0.825 & 0.628 \\
      &                       &                   & \textsc{ESAP}  & 0.858 & \textbf{0.363} & 0.611 & 0.439 & 0.901 & \textbf{0.752} & \textbf{0.827} & \textbf{0.634} \\
      \cmidrule(lr){3-12}
      &                       & \multirow{2}{*}{SEER} & \textsc{Uni}  & 0.851 & 0.363 & 0.607 & 0.449 & 0.891 & 0.636 & 0.764 & 0.561 \\
      &                       &                       & \textsc{ESAP} & \textbf{0.861} & \textbf{0.385} & \textbf{0.623} & \textbf{0.482} & \textbf{0.897} & \textbf{0.716} & \textbf{0.806} & \textbf{0.577} \\
      \cmidrule(lr){3-12}
      &                       & \multirow{2}{*}{Freq} & \textsc{Uni}  & \textbf{0.862} & 0.357 & 0.609 & 0.446 & 0.891 & 0.658 & 0.774 & 0.558 \\
      &                       &                             & \textsc{ESAP} & 0.860 & \textbf{0.396} & \textbf{0.628} & \textbf{0.463} & \textbf{0.907} & \textbf{0.734} & \textbf{0.821} & \textbf{0.572} \\
      \cmidrule(lr){3-12}
      &                       & \multirow{2}{*}{REAP} & \textsc{Uni}   & \textbf{0.872} & \textbf{0.385} & \textbf{0.629} & 0.565 & 0.910 & 0.784 & 0.847 & \textbf{0.706} \\
      &                       &                        & \textsc{ESAP}  & 0.865 & \textbf{0.385} & 0.625 & \textbf{0.588} & \textbf{0.915} & \textbf{0.806} & \textbf{0.861} & 0.704 \\
    \cmidrule(lr){2-12}
      & \multirow{8}{*}{50\%}
      & \multirow{2}{*}{EAN} & \textsc{Uni}   & \textbf{0.846} & 0.341 & \textbf{0.594} & 0.231 & 0.833 & 0.456 & 0.644 & 0.516 \\
      &                       &                   & \textsc{ESAP}  & 0.839 & \textbf{0.346} & 0.593 & \textbf{0.243} & \textbf{0.846} & \textbf{0.494} & \textbf{0.670} & \textbf{0.518} \\
      \cmidrule(lr){3-12}
      &                       & \multirow{2}{*}{SEER} & \textsc{Uni}  & 0.700 & 0.247 & 0.473 & 0.112 & \textbf{0.605} & 0.144 & 0.374 & \textbf{0.455} \\
      &                       &                       & \textsc{ESAP} & \textbf{0.767} & \textbf{0.264} & \textbf{0.516} & \textbf{0.142} & 0.550 & \textbf{0.220} & \textbf{0.385} & 0.446 \\
      \cmidrule(lr){3-12}
      &                       & \multirow{2}{*}{Freq} & \textsc{Uni}  & 0.700 & 0.225 & 0.462 & 0.110 & 0.592 & 0.128 & 0.360 & 0.450 \\
      &                       &                             & \textsc{ESAP} & \textbf{0.781} & \textbf{0.275} & \textbf{0.528} & \textbf{0.182} & \textbf{0.697} & \textbf{0.214} & \textbf{0.455} & \textbf{0.510} \\
      \cmidrule(lr){3-12}
      &                       & \multirow{2}{*}{REAP} & \textsc{Uni}   & 0.828 & \textbf{0.341} & 0.585 & \textbf{0.299} & \textbf{0.872} & \textbf{0.798} & \textbf{0.835} & \textbf{0.596} \\
      &                       &                        & \textsc{ESAP}  & \textbf{0.855} & 0.335 & \textbf{0.595} & 0.267 & 0.867 & 0.792 & 0.830 & 0.585 \\

    \bottomrule
  \end{tabular}%
  }
\end{table*}

\begin{table}[t]
  \centering
  \caption{Search cost and model-loading GPU memory at 25\% global sparsity. We report the GPU configuration, wall-clock search time, and the \textit{bfloat16} GPU memory required to load the pruned and full models. Pruning reduces model-loading memory by roughly 23\%--24\% across models.}
  \label{tab:search_time_memory}
  \resizebox{0.48\textwidth}{!}{%
  \begin{tabular}{l l c S[table-format=3.2] l}
    \toprule
    \textbf{Model} & \textbf{GPU} & \textbf{Generations} & \textbf{Time (h)} & \textbf{Memory (GB)} \\
    \midrule
    OLMoE-7B   & 1$\times$L40S & 50 & 5.8 & \makecell{9.89 / 12.9} \\
    DeepSeek-16B & 1$\times$L40S & 30 & 6.7 & \makecell{22.57 / 29.27} \\
    ERNIE-21B  & 2$\times$L40S & 20 & 5.0 & \makecell{31.17 / 40.66} \\
    Qwen3-30B  & 2$\times$L40S & 10 & 5.2 & \makecell{43.42 / 56.92} \\
    \bottomrule
  \end{tabular}
  }
\end{table}

\begin{table}[t]
  \centering
  \caption{Comparison with DiEP on OLMoE-7B at 25\% global sparsity. Since DiEP jointly optimizes within-layer pruning order and layer-wise sparsity allocation, we include both calibration (0.75h) and search time (1.17h) in EvoESAP’s reported runtime to ensure a fair comparison.}
  \label{tab:diep_comparison}
  \resizebox{0.48\textwidth}{!}{%
  \begin{tabular}{l l c c c c}
    \toprule
    \textbf{Method} & \textbf{GPU} & \textbf{Time (h)} & \textbf{Code Avg} & \textbf{WildBench} & \textbf{MC Avg} \\
    \midrule
    DiEP & 2$\times$L40S & 11.25 & 0.114 & 0.208 & 0.495 \\
    Ours & 1$\times$L40S & \textbf{1.92} & \textbf{0.178} & \textbf{0.275} & \textbf{0.565} \\
    \bottomrule
  \end{tabular}%
  }
  \vspace{-1.6em}
\end{table}

\section{Experiment}
\subsection{Experimental Setup}
\label{sec:implementation_details}
\noindent \textbf{Models and data.} We validate EvoESAP on SMoE LLMs spanning the 7B--30B scale:
OLMoE-1B-7B-0125-Instruct~\cite{muennighoff2024olmoe}, DeepSeek-V2-Lite-Chat~\cite{deepseekv2}, ERNIE-4.5-21B-A3B-PT~\cite{ernie2025technicalreport}, and Qwen3-30B-A3B-Instruct-2507~\cite{yang2025qwen3}. We instantiate within-layer pruning orders using four expert-importance criteria: activation Frequency, SEER soft counting~\cite{muzio2024seer}, Expert Activation Norm (EAN)~\cite{jaiswal2025finding}, and REAP~\cite{lasby2025reap}. In \cref{tab:main_table_olmoe_ernie} and \cref{tab:main_table_deepseek_qwen3}, all pruning metrics are calibrated with 1{,}024 samples from \nolinkurl{evol-codealpaca-v1} and searched with 64 samples from \nolinkurl{tulu-3-sft-personas-math}. Results using C4 as the calibration dataset are reported in \cref{app:c4_results}.

\noindent \textbf{Evaluation.} Following~\cite{lasby2025reap}, our multiple-choice (MC) suite includes ARC-C/ARC-E, BoolQ, HellaSwag, MMLU, OBQA, RTE, and WinoGrande (WinoG.)~\cite{clark2018think,clark2019boolq,zellers2019hellaswag,hendrycks2020measuring,mihaylov2018can,bentivogli2009fifth,sakaguchi2021winogrande}. We evaluate all MC tasks using the standard log-likelihood protocol implemented in \nolinkurl{lm-eval-harness}~\cite{gao2021framework}. For open-ended generation, we evaluate code on EvalPlus and 182 LiveCodeBench problems collected between January and April 2025~\cite{liu2023your,jain2024livecodebench}, math on GSM8K and MATH-500 using \nolinkurl{evalscope}~\cite{cobbe2021training,hendrycks2021measuring,evalscope_2024}, and creative writing on 146 WildBench prompts scored by \nolinkurl{gpt-oss-120b}~\cite{lin2024wildbench,openai2025gptoss120bgptoss20bmodel}. Following the standard protocol, all evaluations are conducted in the zero-shot setting.

\noindent \textbf{Implementation details.} For generation tasks, we use greedy decoding (temperature $=0.0$) for fair comparisons. For Qwen3-30B-A3B, we disable reasoning on all tasks by setting \texttt{enable\_thinking} to \texttt{False} in the chat template. Across all variants, we fix the search hyperparameters: seed 42, population size $P{=}32$, elite size $m{=}4$, maximum transfer $\Delta_{\max}{=}4$, and maximum composed steps $\tau_{\max}{=}3$ (except for the Qwen3+REAP setting, where we use $\tau_{\max}{=}5$). We run the search for $T{=}50$ generations on OLMoE, $T{=}20$ on ERNIE, $T{=}30$ on DeepSeek-V2, and $T{=}10$ on Qwen3. We further restrict each transfer to be even-valued, so that the resulting non-uniform allocations remain compatible with expert parallelism. More generally, for larger models deployed with a larger expert-parallel (EP) world size, this compatibility can be maintained simply by constraining the transfer size $\Delta$ to the corresponding EP granularity.

\subsection{Main Results}
\label{sec:main_results}
\noindent \textbf{Performance comparison}
\Cref{tab:main_table_olmoe_ernie,tab:main_table_deepseek_qwen3} compare \textsc{Uni} (uniform layer-wise sparsity) against the non-uniform allocation found by EvoESAP (reported as \textsc{ESAP}), while \emph{holding the within-layer pruning order fixed} for each criterion (EAN, SEER, Frequency, REAP) and keeping the same global sparsity budget. Across models, EvoESAP most consistently improves open-ended generation, while MC typically changes only slightly; moreover, the gains generally grow as sparsity increases. Detailed per-benchmark results are provided in \cref{app:full_evaluation_results}. On OLMoE at 25\% sparsity, EvoESAP with REAP yields clear generation gains: coding improves by 2.9\%, math by 2.8\%, while MC remains essentially unchanged (+0.2\%). This shows that even under an identical pruning metric, optimizing where capacity is kept can materially strengthen generation-preserving pruning. On DeepSeek at 25\% sparsity, EvoESAP with SEER yields the clearest broad improvement, raising Code Avg by 3.6\%, WildBench by 3.0\%, Math Avg by 1.1\%, and MC by 0.8\%. For the other pruning criteria, the benefits are more concentrated on WildBench, Math and MC, with only modest trade-offs in coding. For ERNIE, improvements are smaller at 25\% (e.g., Frequency: Code Avg +2.5\%, Math Avg +1.4\%, MC +1.1\%), but become much larger at 50\%, where allocation is more consequential for generation. For example, SEER improves Math Avg by +8.7\% with only a small MC change ($-0.4\%$), and Frequency yields an even larger Math Avg gain of +15.2\% (MATH-500 +19.6\%). On Qwen3, EvoESAP most strongly benefits weaker criteria whose \textsc{Uni} allocations leave clear headroom. At 50\% sparsity, Frequency improves Code Avg by 6.6\%, WildBench by 7.2\%, Math Avg by 9.5\%, and MC by 6.0\%. At 25\% sparsity, both SEER and Frequency improve broadly. 
For stronger pruning orders such as REAP on Qwen3 at 25\%, the gains from re-allocation become smaller: the searched allocation improves WildBench and Math Avg (+2.3\% and +1.4\%) while leaving Code Avg and MC essentially unchanged ($-0.4\%$ and $-0.2\%$). This suggests that when the within-layer pruning order already preserves the key experts, there is still room for allocation search to improve performance, but the additional gains are naturally more modest. The tables also underscore that the ``best'' pruning criterion is not universal across models. At 25\% sparsity, REAP is strongest on Qwen3 (uniform REAP achieves the best Code Avg and MC), whereas on OLMoE the same uniform REAP is comparatively weak on coding, even trailing the simpler Frequency ($-2.2\%$ on Code Avg). This non-universality suggests treating allocation as an orthogonal, reusable improvement axis: regardless of which criterion is strongest for a given model, EvoESAP can further optimize where capacity is retained under the same global budget, providing a stable pathway to improve pruning---especially at higher sparsity where preserving open-ended generation quality becomes increasingly sensitive to allocation. Additional visualizations of the searched layer-wise sparsity schedules are provided in \cref{app:visualization_of_searched_sparsity_distribution}. Overall, EvoESAP continues to deliver consistent improvements over uniform allocation under the same global budget.

\begin{table*}[tbp]
  \centering
  \caption{Ablations of fitness choice and search hyperparameters at 25\% global sparsity on OLMoE. We include the uniform REAP baseline as a reference. Blue cells mark the default configuration corresponding to the OLMoE 25\% REAP+\textsc{ESAP} result in \cref{tab:main_table_olmoe_ernie}. Relative to the uniform REAP baseline, all searched settings improve both Code Avg and Math Avg. Among the searched fitness functions, ESAP gives the strongest Code Avg and Math Avg.}
  \label{tab:ablate_fitness_and_samples}
  \resizebox{0.85\linewidth}{!}{%
  \begin{tabular}{l l | c c c | c | c c c | c}
    \toprule
    \multicolumn{2}{c}{} & \multicolumn{3}{c|}{\textbf{Coding}} & \multicolumn{1}{c|}{\textbf{Creative Writing}} & \multicolumn{3}{c|}{\textbf{Math}} & \multicolumn{1}{c}{\textbf{MC}} \\
    \textbf{Ablation} & \textbf{Value} & \textbf{Eval+} & \textbf{LiveCode} & \textbf{Code Avg} & \textbf{WildBench} & \textbf{GSM8K} & \textbf{MATH-500} & \textbf{Math Avg} & \textbf{MC Avg} \\
    \midrule
    Reference & REAP + \textsc{Uni} & 0.314 & 0.005 & 0.160 & 0.292 & 0.596 & 0.200 & 0.398 & 0.579 \\
    \midrule
    \multirow{4}{*}{Fitness} & KL   & 0.331 & 0.016 & 0.174 & 0.285 & 0.595 & 0.216 & 0.405 & 0.582 \\
                             & NLL  & 0.334 & 0.016 & 0.175 & \textbf{0.295} & 0.619 & \textbf{0.230} & 0.424 & 0.576 \\
                             & SAP  & 0.339 & 0.005 & 0.172 & 0.289 & 0.622 & 0.218 & 0.420 & \textbf{0.584} \\
                             & \cellcolor[HTML]{E9F2FB}ESAP (Ours) & \cellcolor[HTML]{E9F2FB}\textbf{0.344} & \cellcolor[HTML]{E9F2FB}\textbf{0.033} & \cellcolor[HTML]{E9F2FB}\textbf{0.189} & \cellcolor[HTML]{E9F2FB}0.279 & \cellcolor[HTML]{E9F2FB}\textbf{0.636} & \cellcolor[HTML]{E9F2FB}0.216 & \cellcolor[HTML]{E9F2FB}\textbf{0.426} & \cellcolor[HTML]{E9F2FB}0.581 \\
    \midrule
    \multirow{4}{*}{Samples} & 8    & 0.320 & 0.022 & 0.171 & \textbf{0.290} & 0.627 & \textbf{0.228} & 0.427 & 0.576 \\
                             & 32   & 0.339 & 0.016 & 0.178 & 0.285 & \textbf{0.645} & 0.222 & \textbf{0.433} & \textbf{0.582} \\
                             & \cellcolor[HTML]{E9F2FB}64   & \cellcolor[HTML]{E9F2FB}\textbf{0.344} & \cellcolor[HTML]{E9F2FB}\textbf{0.033} & \cellcolor[HTML]{E9F2FB}\textbf{0.189} & \cellcolor[HTML]{E9F2FB}0.279 & \cellcolor[HTML]{E9F2FB}0.636 & \cellcolor[HTML]{E9F2FB}0.216 & \cellcolor[HTML]{E9F2FB}0.426 & \cellcolor[HTML]{E9F2FB}0.581 \\
                             & 128  & 0.347 & 0.011 & 0.179 & 0.286 & 0.625 & 0.214 & 0.419 & 0.579 \\
    \midrule
    \multirow{4}{*}{Generations} & 10  & 0.333 & 0.022 & 0.178 & 0.275 & 0.605 & 0.226 & 0.415 & 0.565 \\
                                 & 30  & 0.319 & 0.022 & 0.171 & 0.279 & 0.633 & 0.212 & 0.422 & 0.575 \\
                                 & \cellcolor[HTML]{E9F2FB}50  & \cellcolor[HTML]{E9F2FB}\textbf{0.344} & \cellcolor[HTML]{E9F2FB}\textbf{0.033} & \cellcolor[HTML]{E9F2FB}\textbf{0.189} & \cellcolor[HTML]{E9F2FB}0.279 & \cellcolor[HTML]{E9F2FB}\textbf{0.636} & \cellcolor[HTML]{E9F2FB}0.216 & \cellcolor[HTML]{E9F2FB}0.426 & \cellcolor[HTML]{E9F2FB}0.581 \\
                                 & 70  & 0.342 & 0.011 & 0.177 & \textbf{0.288} & 0.635 & \textbf{0.234} & \textbf{0.434} & \textbf{0.584} \\
    \midrule
    \multirow{4}{*}{Populations} & 8  & 0.318 & 0.027 & 0.173 & 0.280 & 0.634 & 0.196 & 0.415 & 0.582 \\
                                 & 16 & \textbf{0.348} & 0.027 & 0.188 & \textbf{0.285} & \textbf{0.636} & 0.204 & 0.420 & \textbf{0.584} \\
                                 & \cellcolor[HTML]{E9F2FB}32 & \cellcolor[HTML]{E9F2FB}0.344 & \cellcolor[HTML]{E9F2FB}\textbf{0.033} & \cellcolor[HTML]{E9F2FB}\textbf{0.189} & \cellcolor[HTML]{E9F2FB}0.279 & \cellcolor[HTML]{E9F2FB}\textbf{0.636} & \cellcolor[HTML]{E9F2FB}0.216 & \cellcolor[HTML]{E9F2FB}0.426 & \cellcolor[HTML]{E9F2FB}0.581 \\
                                 & 64 & 0.345 & 0.027 & 0.186 & 0.281 & 0.632 & \textbf{0.224} & \textbf{0.428} & 0.583 \\
    \bottomrule
  \end{tabular}%
  }
\end{table*}

\noindent \textbf{Comparison with DiEP.} We further compare EvoESAP with DiEP, a gradient-based framework that jointly optimizes within-layer expert ordering and layer-wise sparsity allocation. We conduct this comparison for OLMoE-1B-7B-0125-Instruct at 25\% global sparsity. For DiEP, we follow the hyperparameter setting reported in the original paper: 10 epochs, learning rate $5\times10^{-3}$, and $\lambda=0.01$. For a fair comparison, both methods use 1024 calibration samples from \textit{evol-codealpaca-v1}. For EvoESAP, we report the result after 10 generations to match DiEP's 10-epoch search budget. As shown in \cref{tab:diep_comparison}, EvoESAP is both more efficient and more effective.

\begin{table}[tbp]
  \centering
  \caption{Comparison of true speculative-decoding acceptance (SPEC-DEC) and the proposed ESAP. ESAP significantly reduces the search time from 29.49 hours to 1.64 hours.}
  \label{tab:specdec_vs_espdataset_code_search}
  \setlength{\tabcolsep}{5pt}
  \small
  \resizebox{0.48\textwidth}{!}{%
  \begin{tabular}{l l c c c c}
    \toprule
    \textbf{Fitness} & \textbf{GPU} & \textbf{Time (h)} &
    \textbf{Code Avg} & \textbf{WildBench} & \textbf{MC Avg} \\
    \midrule
    SPEC-DEC & 2$\times$L40S & 29.49 & 0.171 & \textbf{0.269} & \textbf{0.565} \\
    ESAP     & 1$\times$L40S & \textbf{1.64} & \textbf{0.173} & 0.256 & 0.557 \\
    \bottomrule
  \end{tabular}
  }
\end{table}

\noindent \textbf{Search time and memory usage after compression.}
\cref{tab:search_time_memory} reports the practical cost of running EvoESAP at 25\% global sparsity: the required number of GPUs and the wall-clock search time for the full evolutionary run (with the model-specific generation counts shown in the table). We also report the peak GPU memory required to load the \emph{final} compressed model for inference, alongside the full model under the same \texttt{bfloat16} setting. Across models, the pruned checkpoints show roughly 25\% lower GPU memory usage than the full models, reflecting the primary deployment benefit of whole-expert pruning. For large SMoE models, the activated parameters per token are already much smaller than the full model size, so deployment is often constrained more by GPU memory than by per-token latency. For example, Qwen3-30B-A3B activates about 3B parameters per token at inference, but serving it still requires loading the full 30B-parameter model. Accordingly, whole-expert pruning is primarily a memory-reduction approach; methods that target latency gains usually reduce computation within activated experts (e.g., intra-expert compression), which is orthogonal to whole-expert pruning and outside the scope of this paper.
\subsection{Ablation Study}
All experiments in this section use OLMoE-1B-7B-0125-Instruct and adopt REAP calibrated on \texttt{evol-codealpaca-v1}, as the within-layer pruning order. Unless otherwise specified, we follow the same experimental settings as in \cref{sec:implementation_details}.

\noindent \textbf{Speculative decoding as fitness.} To validate ESAP as a proxy for speculative-decoding compatibility, we additionally run evolutionary search using the \emph{true} speculative-decoding acceptance (SPEC-DEC) as the fitness. We summarize the comparison in \cref{tab:specdec_vs_espdataset_code_search}, where we run the search on \texttt{evol-codealpaca-v1} for 30 generations. Directly optimizing SPEC-DEC yields only small downstream differences relative to ESAP, yet it is impractical as an inner-loop fitness: it requires running both the draft and verifier with large KV caches and performing autoregressive decoding, substantially increasing GPU demand and wall-clock evaluation cost. In contrast, ESAP achieves comparable performance while using fewer GPUs and reducing search time by \(\sim\!18\times\), supporting ESAP as an effective proxy for speculative-decoding compatibility.

\noindent \textbf{Comparison with other teacher-forcing fitness functions.} To examine the effectiveness of our proposed ESAP, we compare it against three alternative fitness signals: Kullback--Leibler divergence (KL), negative log-likelihood (NLL), and speculative acceptance proxy (SAP). Results are reported in \cref{tab:ablate_fitness_and_samples}, where we also include the uniform REAP model as a reference. All searched fitness variants improve over this uniform baseline on both Code Avg and Math Avg, indicating that the benefit is not tied to a single search objective. Among them, ESAP provides the strongest and most consistent generation-preserving compression: it attains the best \emph{coding} performance across Eval+, LiveCode, and Code Avg, and also achieves the highest \emph{overall math} score with the strongest GSM8K. Importantly, ESAP remains competitive on multiple-choice evaluation, coming close to the best MC score. This suggests ESAP is a promising fitness for generation-preserving sparsity allocation.

\noindent \textbf{Sensitivity to the amount of search data.}
\cref{tab:ablate_fitness_and_samples} also examines the effect of search sample size. All searched sample-size settings remain above the uniform REAP reference on both Code Avg and Math Avg, showing that the benefit of allocation search is robust even with a small search set. Overall, once we use a modest number of samples, search quality is largely stable and does not improve monotonically with more data.
In particular, 32--64 samples already capture the benefit of EvoESAP: 64 samples deliver the strongest coding results,
while 32 samples achieve the best Math Avg and MC Avg, and also attain the highest GSM8K score.
With only 8 samples, performance remains competitive but is slightly weaker, consistent with a noisier fitness estimate.
By contrast, increasing to 128 samples provides no additional gains. We therefore adopt 64 samples as our default experimental setting.

\noindent \textbf{Sensitivity to search generations and population size.}
\cref{tab:ablate_fitness_and_samples} further reports ablations on the number of search generations and the population size. Again, every searched setting outperforms the uniform REAP reference on both Code Avg and Math Avg, suggesting that the gains do not depend on a narrowly tuned search budget. Across generation counts, performance remains fairly stable while benefiting from a longer search budget overall: 50 generations yield the strongest coding performance and the best GSM8K score, while 70 generations slightly lead on WildBench, Math Avg, and MC Avg. Varying the population size shows a similar tradeoff. A population size of 32 gives the best Code Avg while tying for the best GSM8K score, 16 gives the strongest Eval+, WildBench, and MC Avg, and 64 slightly improves Math Avg. Overall, these results indicate that EvoESAP is robust to hyperparameter choices, and 50 generations with a population size of 32 is a reasonable default.

\section{Conclusion}
We decouple SMoE expert pruning into within-layer expert ranking and across-layer budget allocation, and propose EvoESAP, a plug-and-play framework that addresses the underexplored problem of non-uniform layer-wise sparsity allocation under a fixed global budget. Given any within-layer pruning order (e.g., Frequency, SEER, EAN, or REAP), EvoESAP searches over integer layer-wise budgets via a budget-preserving level-switch mutation, guided by Expected Speculative Acceptance Proxy (ESAP), a speculative-decoding-inspired, teacher-forced overlap score between the baseline and pruned next-token distributions. Across 7B--30B SMoE models and 25\%--50\% sparsity, the non-uniform schedules discovered by \textsc{EvoESAP} consistently improve downstream performance over uniform pruning under the same pruning metric and global budget, with the most reliable gains on open-ended generation and with particularly large benefits at higher sparsity levels.

\bibliographystyle{ACM-Reference-Format}
\bibliography{sample-base}

\clearpage
\onecolumn
\appendix
\appendix

\section{Evolutionary Search Pseudocode}
\label[appendix]{app:evo_alg}

\begin{algorithm}[H]
\caption{Evolutionary Search for Non-uniform Layer-wise Sparsity Allocation}
\label{alg:level_switch_es}
\begin{algorithmic}[1]
\REQUIRE Per-layer pruning orders $\{\pi_\ell\}_{\ell=1}^{L}$; expert counts $\{n_\ell\}$; fanouts $\{k_\ell\}$;
global budget $B$; search dataset $\mathcal{D}_{\text{search}}=\{(u^{(i)},a^{(i)})\}_{i=1}^{N}$;
fitness $f(\mathbf{r};\mathcal{D}_{\text{search}})$ (ESAP);
population size $P$; elite size $m$; generations $T$; max transfer $\Delta_{\max}$; mutation cap $\tau_{\max}$.
\ENSURE Best allocation $\mathbf{r}^\star$.

\STATE \textbf{Feasible set:}
$\mathcal{F} \triangleq \Big\{\mathbf{r}\in\mathbb{Z}^L \;\big|\; \sum_{\ell=1}^{L} r_\ell = B,\;\; 0\le r_\ell \le n_\ell-k_\ell\;\;\forall \ell \Big\}$.

\STATE \textbf{Init population} $\mathcal{S}^{(0)} \subset \mathcal{F}$ of size $P$ using (i) uniform, (ii) patterned seeds, and (iii) random feasible samples.

\STATE \textbf{(Optional speedup)} Cache baseline (full-model) logits on $\mathcal{D}_{\text{search}}$ under teacher forcing for reuse in $f(\cdot;\mathcal{D}_{\text{search}})$.

\STATE Evaluate $f(\mathbf{r};\mathcal{D}_{\text{search}})$ for all $\mathbf{r}\in\mathcal{S}^{(0)}$.
\STATE $\mathbf{r}^\star \leftarrow \arg\max_{\mathbf{r}\in \mathcal{S}^{(0)}} f(\mathbf{r};\mathcal{D}_{\text{search}})$.

\FOR{$t = 0,1,\dots,T-1$}
  \STATE \textbf{Selection:} $\mathcal{S}^{(t)}_{\mathrm{elite}} \leftarrow \mathrm{TopM}\!\left(\mathcal{S}^{(t)};\ f(\cdot;\mathcal{D}_{\text{search}}),\ m\right)$.
  \STATE $\mathcal{S}^{(t+1)} \leftarrow \mathcal{S}^{(t)}_{\mathrm{elite}}$.

  \WHILE{$|\mathcal{S}^{(t+1)}| < P$}
    \STATE Sample parent $\mathbf{r}$ uniformly from $\mathcal{S}^{(t)}_{\mathrm{elite}}$.
    \STATE Sample mutation count
    \[
      \tau \leftarrow \min\big(U\{1,\dots,\tau_{\max}\},\; U\{1,\dots,\tau_{\max}\}\big).
    \]
    \STATE $\mathbf{r}' \leftarrow \mathbf{r}$.
    \FOR{$i = 1,2,\dots,\tau$}
      \REPEAT
        \STATE Sample distinct layers $a\neq b$ uniformly from $\{1,\dots,L\}$.
        \STATE Sample $\Delta$ uniformly from $\{1,2,\dots,\Delta_{\max}\}$.
        \STATE Propose level-switch update:
        \[
        \tilde{r}_\ell =
        \begin{cases}
        r'_\ell + \Delta, & \ell=a,\\
        r'_\ell - \Delta, & \ell=b,\\
        r'_\ell, & \text{otherwise}.
        \end{cases}
        \]
      \UNTIL{$\tilde{\mathbf{r}} \in \mathcal{F}$}
      \STATE $\mathbf{r}' \leftarrow \tilde{\mathbf{r}}$.
    \ENDFOR
    \STATE $\mathcal{S}^{(t+1)} \leftarrow \mathcal{S}^{(t+1)} \cup \{\mathbf{r}'\}$.
  \ENDWHILE

  \STATE Evaluate $f(\mathbf{r};\mathcal{D}_{\text{search}})$ for all newly added $\mathbf{r}\in\mathcal{S}^{(t+1)}$ if not yet evaluated.
  \STATE $\mathbf{r}^\star \leftarrow \arg\max_{\mathbf{r}\in \{\mathbf{r}^\star\}\cup\mathcal{S}^{(t+1)}} f(\mathbf{r};\mathcal{D}_{\text{search}})$.
\ENDFOR

\RETURN $\mathbf{r}^\star$.
\end{algorithmic}
\end{algorithm}

\section{Derivation of the Total-Variation Relation}
\label[appendix]{app:esap_tv}

In this appendix we derive the identity
\begin{equation}
\mathrm{ESAP}(x) \;=\; 1 - \mathrm{TV}\!\left(p(\cdot\mid x),\,q(\cdot\mid x)\right),
\end{equation}
where $p(\cdot\mid x)$ and $q(\cdot\mid x)$ are categorical distributions over the vocabulary $\mathcal{V}$, and
\begin{equation}
\mathrm{TV}(p,q)\;\triangleq\;\tfrac12 \sum_{v\in\mathcal{V}} \big|p(v\mid x)-q(v\mid x)\big|.
\end{equation}

Recall the closed form of ESAP:
\begin{equation}
\mathrm{ESAP}(x)
\;=\;
\sum_{v\in\mathcal{V}} \min\!\big(p(v\mid x),\,q(v\mid x)\big).
\label{eq:esap_closed_app}
\end{equation}
We use the elementary identity valid for any $a,b\ge 0$:
\begin{equation}
\min(a,b) \;=\; \tfrac12\big(a+b-|a-b|\big).
\label{eq:min_identity_app}
\end{equation}
Applying \eqref{eq:min_identity_app} pointwise to each token $v$ and summing gives
\begin{align}
\sum_{v\in\mathcal{V}} \min\!\big(p(v\mid x),q(v\mid x)\big)
&= \tfrac12 \sum_{v\in\mathcal{V}} \Big(p(v\mid x)+q(v\mid x)-|p(v\mid x)-q(v\mid x)|\Big) \nonumber\\
&= \tfrac12 \Bigg(\sum_{v\in\mathcal{V}} p(v\mid x) + \sum_{v\in\mathcal{V}} q(v\mid x)\Bigg)
     - \tfrac12 \sum_{v\in\mathcal{V}} |p(v\mid x)-q(v\mid x)| \nonumber\\
&= 1 - \tfrac12 \sum_{v\in\mathcal{V}} |p(v\mid x)-q(v\mid x)|,
\label{eq:esap_to_l1_app}
\end{align}
where we used $\sum_{v} p(v\mid x)=\sum_{v} q(v\mid x)=1$. Recognizing the last term as the total-variation distance yields
\begin{equation}
\mathrm{ESAP}(x)
\;=\;
1 - \mathrm{TV}\!\left(p(\cdot\mid x),\,q(\cdot\mid x)\right).
\end{equation}

\section{Results using C4 as calibration dataset}
\label[appendix]{app:c4_results}

\begin{table*}[tbp]
  \centering
  \caption{MC and open-ended generation benchmark results for ERNIE-4.5-21B-A3B-PT under 25\% global sparsity. Pruning is calibrated on C4. \textsc{ESAP} is the non-uniform allocation searched on \texttt{evol-codealpaca-v1} with the same pruning metric.}
  \label{tab:ernie_c4_025}
  \resizebox{0.8\textwidth}{!}{%
  \begin{tabular}{l l l l | c c c | c | c c c | c}
    \toprule
    \multicolumn{4}{c}{} & \multicolumn{3}{c|}{\textbf{Coding}} & \multicolumn{1}{c|}{\textbf{Creative Writing}} & \multicolumn{3}{c|}{\textbf{Math}} & \multicolumn{1}{c}{\textbf{MC}} \\
    \textbf{Model} & \textbf{Sparsity} & \textbf{Method} &  &
    \textbf{Eval+} & \textbf{LiveCode} & \textbf{Avg} &
    \textbf{WildBench} &
    \textbf{GSM8K} & \textbf{MATH-500} & \textbf{Avg} &
    \textbf{MC Avg} \\
    \midrule

    \multirow{10}{*}{\makecell[l]{ERNIE}}
      & \multicolumn{3}{l|}{\textbf{Full}} & 0.867 & 0.247 & 0.557 & 0.479 & 0.829 & 0.780 & 0.804 & 0.721 \\
    \cmidrule(lr){2-12}
      & \multirow{8}{*}{25\%}

      & \multirow{2}{*}{EAN}  & \textsc{Uni}  & 0.248 & 0.049 & 0.148 & 0.412 & 0.670 & 0.366 & 0.518 & \textbf{0.685} \\
      &                       & & \textsc{ESAP} & \textbf{0.354} & \textbf{0.093} & \textbf{0.223} & \textbf{0.433} & \textbf{0.752} & \textbf{0.438} & \textbf{0.595} & 0.669 \\
      \cmidrule(lr){3-12}

      &                       & \multirow{2}{*}{SEER} & \textsc{Uni}  & 0.249 & 0.055 & 0.152 & 0.406 & \textbf{0.738} & 0.374 & 0.556 & 0.660 \\
      &                       &                        & \textsc{ESAP} & \textbf{0.367} & \textbf{0.060} & \textbf{0.213} & \textbf{0.419} & 0.737 & \textbf{0.378} & \textbf{0.557} & \textbf{0.671} \\
      \cmidrule(lr){3-12}

      &                       & \multirow{2}{*}{Freq} & \textsc{Uni}  & 0.269 & \textbf{0.071} & 0.170 & 0.355 & 0.636 & 0.304 & 0.470 & 0.655 \\
      &                       &                        & \textsc{ESAP} & \textbf{0.329} & 0.049 & \textbf{0.189} & \textbf{0.367} & \textbf{0.685} & \textbf{0.418} & \textbf{0.551} & \textbf{0.662} \\
      \cmidrule(lr){3-12}

      &                       & \multirow{2}{*}{REAP} & \textsc{Uni}  & 0.210 & 0.060 & 0.135 & 0.413 & 0.782 & 0.482 & 0.632 & 0.684 \\
      &                       &                        & \textsc{ESAP} & \textbf{0.401} & \textbf{0.132} & \textbf{0.267} & \textbf{0.435} & \textbf{0.817} & \textbf{0.616} & \textbf{0.716} & \textbf{0.692} \\

    \bottomrule
  \end{tabular}%
  }
\end{table*}

We also report results using C4, a cleaned Common Crawl web-text corpus, as the calibration dataset in \cref{tab:ernie_c4_025}. All hyperparameters are the same as those in \cref{sec:implementation_details}. Here we highlight two observations: (i) calibrating on C4 leads to a substantial drop on coding benchmarks compared to \cref{tab:main_table_olmoe_ernie} (calibrated on \texttt{evol-codealpaca-v1}); and (ii) under the same C4 calibration setting, \textsc{EvoESAP} consistently improves performance.

\section{Visualization of searched sparsity distribution}
\label[appendix]{app:visualization_of_searched_sparsity_distribution}
\begin{figure}[tbp]
    \centering
    \begin{subfigure}[t]{\linewidth}
        \centering
        \IfFileExists{figures/density_distribution_searched_ratio_0_250.pdf}{\includegraphics[width=\linewidth]{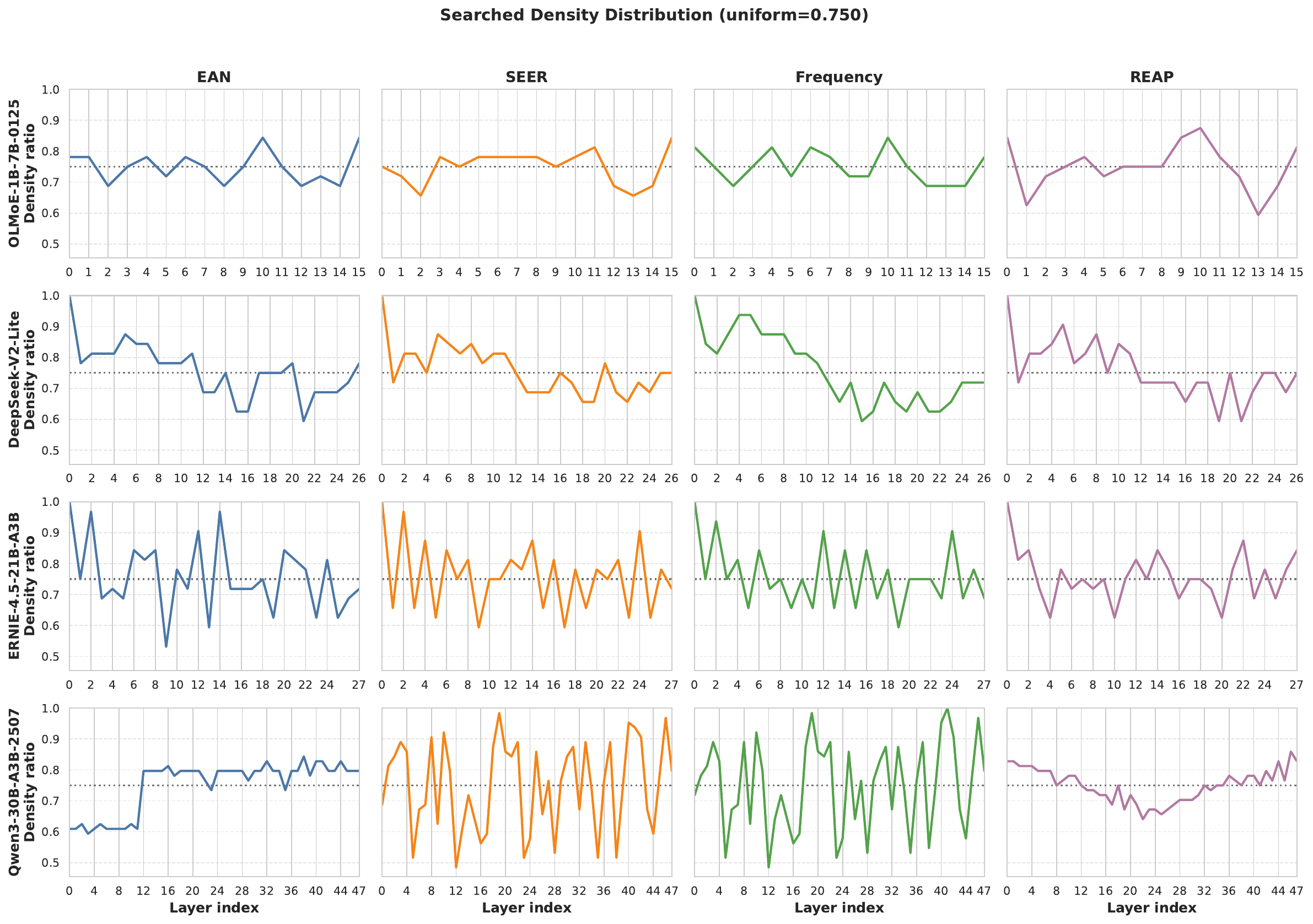}}{\fbox{\parbox{0.9\linewidth}{\centering Missing figure asset:\\ figures/density_distribution_searched_ratio_0_250.pdf}}}
        \caption{25\% global sparsity}
        \label{fig:density_distribution_searched_025}
    \end{subfigure}

    \begin{subfigure}[t]{\linewidth}
        \centering
        \includegraphics[width=\linewidth]{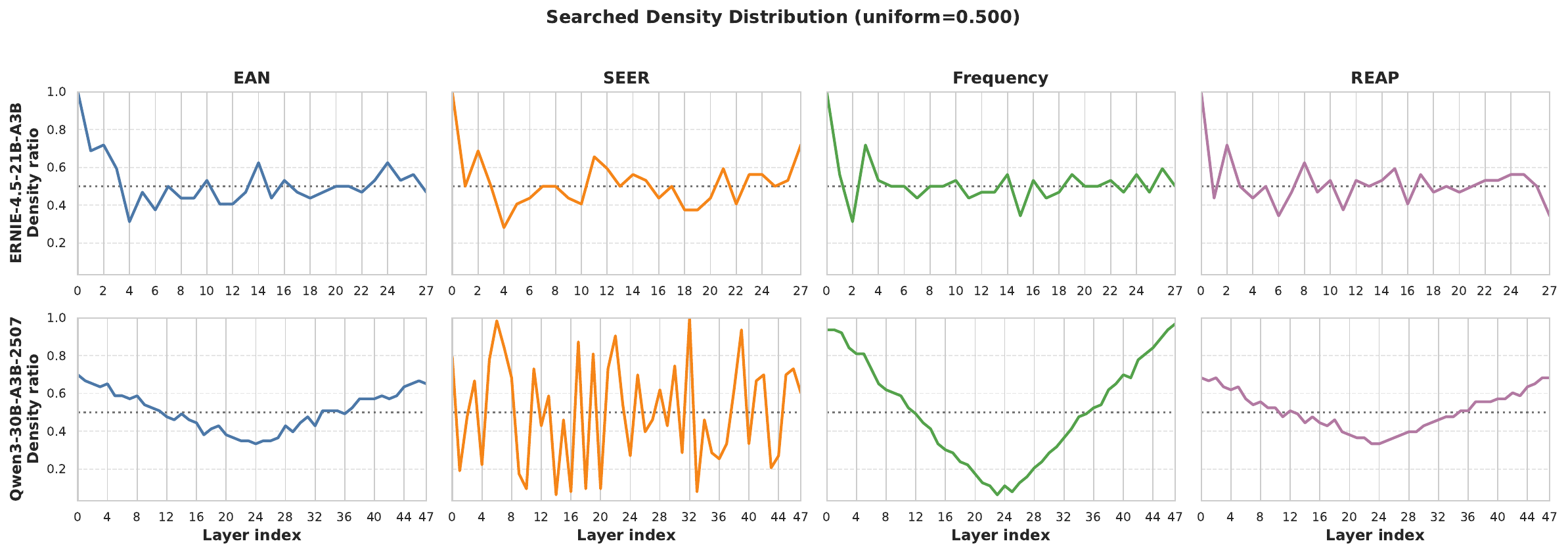}
        \caption{50\% global sparsity}
        \label{fig:density_distribution_searched_050}
    \end{subfigure}
    \caption{Layer-wise density distributions (density $=1-\text{sparsity}$) of the searched non-uniform allocations across different pruning metrics.}
    \Description{Two vertically stacked subfigures summarize searched layer-wise density schedules across pruning metrics. The 25 percent panel uses a placeholder if the local figure asset is unavailable, and the 50 percent panel shows the available searched density visualization.}
    \label{fig:density_distribution_searched}
\end{figure}

\Cref{fig:density_distribution_searched} visualizes the \emph{searched} layer-wise density schedules (density $=1-\text{sparsity}$) produced by EvoESAP under four fixed within-layer pruning orders (EAN, SEER, Frequency, REAP) at two global sparsity levels. We show 25\% sparsity in \cref{fig:density_distribution_searched_025} and 50\% sparsity in \cref{fig:density_distribution_searched_050}, using the same experimental settings as \cref{sec:implementation_details}. For clarity, we plot density rather than sparsity, since higher density directly reflects more retained capacity in each layer. Overall, the searched allocations are non-uniform, but their shapes are not consistent across pruning criteria or backbones, suggesting that there is no single universal allocation template. A particularly illustrative contrast appears at 50\% sparsity: for ERNIE, the searched schedules across different pruning orders are relatively similar and fluctuate around the uniform schedule, whereas for Qwen3 they vary substantially across criteria, with SEER inducing the largest deviations. These observations indicate that optimizing within-layer pruning orders alone is generally insufficient: the allocation that best preserves behavior can depend on both the model and the pruning criterion, making the discovery of effective non-uniform schedules non-trivial.

\section{Full evaluation results}
\label[appendix]{app:full_evaluation_results}

\begin{table*}[tbp]
  \centering
  \caption{Full results on coding benchmarks. Eval+ is the average of HumanEval, HumanEval+, MBPP and MBPP+.}
  \label{tab:appendix_code_full}
  \resizebox{0.98\textwidth}{!}{%
  \begin{tabular}{l l l l | c c c c c c c}
    \toprule
    \multicolumn{4}{c}{} & \multicolumn{7}{c}{\textbf{Coding}} \\
    \textbf{Model} & \textbf{Compression} & \textbf{Method} &  & \textbf{HumanEval} & \textbf{HumanEval$+$} & \textbf{MBPP} & \textbf{MBPP$+$} & \textbf{Eval+} & \textbf{LiveCodeBench} & \textbf{Code Avg} \\
    \midrule
    \multirow{10}{*}{\makecell[l]{OLMoE-1B-7B-\\0125-Instruct}} & \multicolumn{3}{l|}{\textbf{Baseline}} & 0.354 & 0.323 & 0.373 & 0.312 & 0.341 & 0.033 & 0.187 \\
    \cmidrule(lr){2-11}
     & \multirow{8}{*}{25\%} & \multirow{2}{*}{EAN} & \textsc{Uniform}  & \textbf{0.366} & \textbf{0.341} & \textbf{0.360} & \textbf{0.307} & \textbf{0.343} & 0.022 & \textbf{0.183} \\
     &  &  & \textsc{Searched} & 0.317 & 0.293 & 0.347 & 0.291 & 0.312 & \textbf{0.027} & 0.170 \\
    \cmidrule(lr){3-11}
     &  & \multirow{2}{*}{SEER} & \textsc{Uniform}  & \textbf{0.360} & \textbf{0.341} & \textbf{0.349} & \textbf{0.307} & \textbf{0.339} & \textbf{0.027} & \textbf{0.183} \\
     &  &  & \textsc{Searched} & 0.311 & 0.274 & 0.347 & 0.294 & 0.306 & 0.022 & 0.164 \\
    \cmidrule(lr){3-11}
     &  & \multirow{2}{*}{Frequency} & \textsc{Uniform}  & \textbf{0.378} & \textbf{0.335} & 0.349 & 0.302 & 0.341 & \textbf{0.022} & \textbf{0.182} \\
     &  &  & \textsc{Searched} & 0.360 & 0.311 & \textbf{0.376} & \textbf{0.320} & \textbf{0.342} & \textbf{0.022} & \textbf{0.182} \\
    \cmidrule(lr){3-11}
     &  & \multirow{2}{*}{REAP} & \textsc{Uniform}  & 0.341 & 0.311 & 0.331 & 0.272 & 0.314 & 0.005 & 0.160 \\
     &  &  & \textsc{Searched} & \textbf{0.378} & \textbf{0.341} & \textbf{0.352} & \textbf{0.304} & \textbf{0.344} & \textbf{0.033} & \textbf{0.189} \\
    \midrule[\heavyrulewidth]
    \multirow{10}{*}{\makecell[l]{DeepSeek-V2-Lite-\\Chat}} & \multicolumn{3}{l|}{\textbf{Baseline}} & 0.591 & 0.524 & 0.585 & 0.497 & 0.549 & 0.104 & 0.327 \\
    \cmidrule(lr){2-11}
     & \multirow{8}{*}{25\%} & \multirow{2}{*}{EAN} & \textsc{Uniform}  & \textbf{0.537} & \textbf{0.463} & 0.450 & \textbf{0.386} & \textbf{0.459} & \textbf{0.071} & \textbf{0.265} \\
     &  &  & \textsc{Searched} & 0.463 & 0.378 & \textbf{0.458} & 0.368 & 0.417 & \textbf{0.071} & 0.244 \\
    \cmidrule(lr){3-11}
     &  & \multirow{2}{*}{SEER} & \textsc{Uniform}  & 0.409 & 0.348 & 0.460 & 0.399 & 0.404 & 0.099 & 0.252 \\
     &  &  & \textsc{Searched} & \textbf{0.476} & \textbf{0.421} & \textbf{0.511} & \textbf{0.431} & \textbf{0.460} & \textbf{0.115} & \textbf{0.288} \\
    \cmidrule(lr){3-11}
     &  & \multirow{2}{*}{Frequency} & \textsc{Uniform}  & \textbf{0.299} & \textbf{0.274} & \textbf{0.336} & 0.278 & \textbf{0.297} & \textbf{0.099} & \textbf{0.198} \\
     &  &  & \textsc{Searched} & 0.287 & 0.256 & \textbf{0.336} & \textbf{0.291} & 0.292 & 0.082 & 0.187 \\
    \cmidrule(lr){3-11}
     &  & \multirow{2}{*}{REAP} & \textsc{Uniform}  & \textbf{0.518} & \textbf{0.451} & \textbf{0.500} & \textbf{0.421} & \textbf{0.473} & \textbf{0.077} & \textbf{0.275} \\
     &  &  & \textsc{Searched} & 0.463 & 0.427 & 0.492 & 0.386 & 0.442 & 0.071 & 0.257 \\
    \midrule[\heavyrulewidth]
    \multirow{22}{*}{\makecell[l]{ERNIE-4.5-\\21B-A3B-PT}} & \multicolumn{3}{l|}{\textbf{Baseline}} & 0.909 & 0.878 & 0.915 & 0.765 & 0.867 & 0.247 & 0.557 \\
    \cmidrule(lr){2-11}
     & \multirow{8}{*}{25\%} & \multirow{2}{*}{EAN} & \textsc{Uniform}  & 0.884 & 0.854 & 0.844 & \textbf{0.728} & 0.827 & 0.214 & 0.520 \\
     &  &  & \textsc{Searched} & \textbf{0.902} & \textbf{0.866} & \textbf{0.847} & 0.714 & \textbf{0.832} & \textbf{0.225} & \textbf{0.528} \\
    \cmidrule(lr){3-11}
     &  & \multirow{2}{*}{SEER} & \textsc{Uniform}  & 0.890 & 0.860 & 0.844 & \textbf{0.725} & 0.830 & \textbf{0.214} & \textbf{0.522} \\
     &  &  & \textsc{Searched} & \textbf{0.909} & \textbf{0.866} & \textbf{0.852} & \textbf{0.725} & \textbf{0.838} & 0.203 & 0.520 \\
    \cmidrule(lr){3-11}
     &  & \multirow{2}{*}{Frequency} & \textsc{Uniform}  & \textbf{0.878} & \textbf{0.841} & \textbf{0.847} & \textbf{0.706} & \textbf{0.818} & 0.181 & 0.499 \\
     &  &  & \textsc{Searched} & 0.866 & 0.835 & 0.839 & 0.704 & 0.811 & \textbf{0.236} & \textbf{0.524} \\
    \cmidrule(lr){3-11}
     &  & \multirow{2}{*}{REAP} & \textsc{Uniform}  & 0.872 & 0.841 & 0.860 & 0.717 & 0.823 & \textbf{0.231} & \textbf{0.527} \\
     &  &  & \textsc{Searched} & \textbf{0.896} & \textbf{0.848} & \textbf{0.865} & \textbf{0.722} & \textbf{0.833} & 0.209 & 0.521 \\
    \cmidrule(lr){2-11}
     & \multirow{8}{*}{50\%} & \multirow{2}{*}{EAN} & \textsc{Uniform}  & 0.646 & 0.616 & 0.701 & 0.579 & 0.636 & \textbf{0.148} & 0.392 \\
     &  &  & \textsc{Searched} & \textbf{0.677} & \textbf{0.646} & \textbf{0.714} & \textbf{0.601} & \textbf{0.659} & 0.143 & \textbf{0.401} \\
    \cmidrule(lr){3-11}
     &  & \multirow{2}{*}{SEER} & \textsc{Uniform}  & 0.713 & 0.689 & \textbf{0.759} & \textbf{0.630} & 0.698 & 0.170 & 0.434 \\
     &  &  & \textsc{Searched} & \textbf{0.762} & \textbf{0.726} & 0.730 & 0.619 & \textbf{0.709} & \textbf{0.181} & \textbf{0.445} \\
    \cmidrule(lr){3-11}
     &  & \multirow{2}{*}{Frequency} & \textsc{Uniform}  & 0.677 & 0.634 & \textbf{0.704} & \textbf{0.574} & \textbf{0.647} & \textbf{0.143} & \textbf{0.395} \\
     &  &  & \textsc{Searched} & \textbf{0.689} & \textbf{0.640} & 0.690 & 0.569 & \textbf{0.647} & 0.126 & 0.387 \\
    \cmidrule(lr){3-11}
     &  & \multirow{2}{*}{REAP} & \textsc{Uniform}  & 0.768 & 0.726 & \textbf{0.770} & 0.656 & 0.730 & \textbf{0.192} & 0.461 \\
     &  &  & \textsc{Searched} & \textbf{0.774} & \textbf{0.738} & 0.765 & \textbf{0.669} & \textbf{0.737} & 0.187 & \textbf{0.462} \\
    \midrule[\heavyrulewidth]
    \multirow{22}{*}{\makecell[l]{Qwen3-30B-\\A3B-Instruct-\\2507}} & \multicolumn{3}{l|}{\textbf{Baseline}} & 0.939 & 0.902 & 0.892 & 0.751 & 0.871 & 0.368 & 0.619 \\
    \cmidrule(lr){2-11}
     & \multirow{8}{*}{25\%} & \multirow{2}{*}{EAN} & \textsc{Uniform}  & \textbf{0.945} & \textbf{0.896} & \textbf{0.886} & \textbf{0.757} & \textbf{0.871} & \textbf{0.363} & \textbf{0.617} \\
     &  &  & \textsc{Searched} & 0.927 & 0.890 & 0.878 & 0.738 & 0.858 & \textbf{0.363} & 0.611 \\
    \cmidrule(lr){3-11}
     &  & \multirow{2}{*}{SEER} & \textsc{Uniform}  & 0.902 & 0.854 & 0.894 & 0.754 & 0.851 & 0.363 & 0.607 \\
     &  &  & \textsc{Searched} & \textbf{0.915} & \textbf{0.872} & \textbf{0.897} & \textbf{0.759} & \textbf{0.861} & \textbf{0.385} & \textbf{0.623} \\
    \cmidrule(lr){3-11}
     &  & \multirow{2}{*}{Frequency} & \textsc{Uniform}  & \textbf{0.921} & \textbf{0.878} & \textbf{0.897} & 0.751 & \textbf{0.862} & 0.357 & 0.609 \\
     &  &  & \textsc{Searched} & 0.915 & 0.872 & \textbf{0.897} & \textbf{0.757} & 0.860 & \textbf{0.396} & \textbf{0.628} \\
    \cmidrule(lr){3-11}
     &  & \multirow{2}{*}{REAP} & \textsc{Uniform}  & \textbf{0.945} & \textbf{0.896} & \textbf{0.897} & \textbf{0.749} & \textbf{0.872} & \textbf{0.385} & \textbf{0.629} \\
     &  &  & \textsc{Searched} & 0.933 & 0.890 & 0.892 & 0.746 & 0.865 & \textbf{0.385} & 0.625 \\
    \cmidrule(lr){2-11}
     & \multirow{8}{*}{50\%} & \multirow{2}{*}{EAN} & \textsc{Uniform}  & \textbf{0.915} & \textbf{0.878} & \textbf{0.865} & \textbf{0.725} & \textbf{0.846} & 0.341 & \textbf{0.594} \\
     &  &  & \textsc{Searched} & \textbf{0.915} & 0.866 & 0.860 & 0.714 & 0.839 & \textbf{0.346} & 0.593 \\
    \cmidrule(lr){3-11}
     &  & \multirow{2}{*}{SEER} & \textsc{Uniform}  & 0.774 & 0.713 & 0.720 & 0.593 & 0.700 & 0.247 & 0.473 \\
     &  &  & \textsc{Searched} & \textbf{0.872} & \textbf{0.817} & \textbf{0.757} & \textbf{0.622} & \textbf{0.767} & \textbf{0.264} & \textbf{0.516} \\
    \cmidrule(lr){3-11}
     &  & \multirow{2}{*}{Frequency} & \textsc{Uniform}  & 0.787 & 0.738 & 0.701 & 0.574 & 0.700 & 0.225 & 0.462 \\
     &  &  & \textsc{Searched} & \textbf{0.872} & \textbf{0.817} & \textbf{0.788} & \textbf{0.648} & \textbf{0.781} & \textbf{0.275} & \textbf{0.528} \\
    \cmidrule(lr){3-11}
     &  & \multirow{2}{*}{REAP} & \textsc{Uniform}  & 0.902 & 0.854 & 0.857 & 0.698 & 0.828 & \textbf{0.341} & 0.585 \\
     &  &  & \textsc{Searched} & \textbf{0.939} & \textbf{0.902} & \textbf{0.868} & \textbf{0.709} & \textbf{0.855} & 0.335 & \textbf{0.595} \\
    \bottomrule
  \end{tabular}%
  }
\end{table*}

\begin{table*}[tbp]
  \centering
  \caption{Full results on WildBench and math benchmarks.}
  \label{tab:appendix_wild_math_full}
  \resizebox{0.72\textwidth}{!}{%
  \begin{tabular}{l l l l | c c c c}
    \toprule
    \multicolumn{4}{c}{} & \multicolumn{4}{c}{\textbf{WildBench + Math}} \\
    \textbf{Model} & \textbf{Compression} & \textbf{Method} &  & \textbf{WildBench} & \textbf{GSM8K} & \textbf{MATH-500} & \textbf{Math Avg} \\
    \midrule
    \multirow{10}{*}{\makecell[l]{OLMoE-1B-7B-\\0125-Instruct}} & \multicolumn{3}{l|}{\textbf{Baseline}} & 0.444 & 0.682 & 0.222 & 0.452 \\
    \cmidrule(lr){2-8}
     & \multirow{8}{*}{25\%} & \multirow{2}{*}{EAN} & \textsc{Uniform}  & \textbf{0.269} & \textbf{0.585} & 0.190 & 0.387 \\
     &  &  & \textsc{Searched} & 0.258 & 0.576 & \textbf{0.232} & \textbf{0.404} \\
    \cmidrule(lr){3-8}
     &  & \multirow{2}{*}{SEER} & \textsc{Uniform}  & 0.253 & 0.577 & 0.204 & 0.390 \\
     &  &  & \textsc{Searched} & \textbf{0.254} & \textbf{0.601} & \textbf{0.248} & \textbf{0.424} \\
    \cmidrule(lr){3-8}
     &  & \multirow{2}{*}{Frequency} & \textsc{Uniform}  & \textbf{0.265} & 0.591 & \textbf{0.220} & \textbf{0.405} \\
     &  &  & \textsc{Searched} & 0.244 & \textbf{0.596} & 0.208 & 0.402 \\
    \cmidrule(lr){3-8}
     &  & \multirow{2}{*}{REAP} & \textsc{Uniform}  & \textbf{0.292} & 0.596 & 0.200 & 0.398 \\
     &  &  & \textsc{Searched} & 0.279 & \textbf{0.636} & \textbf{0.216} & \textbf{0.426} \\
    \midrule[\heavyrulewidth]
    \multirow{10}{*}{\makecell[l]{DeepSeek-V2-Lite-\\Chat}} & \multicolumn{3}{l|}{\textbf{Baseline}} & 0.418 & 0.610 & 0.298 & 0.454 \\
    \cmidrule(lr){2-8}
     & \multirow{8}{*}{25\%} & \multirow{2}{*}{EAN} & \textsc{Uniform}  & 0.259 & 0.531 & \textbf{0.224} & 0.378 \\
     &  &  & \textsc{Searched} & \textbf{0.301} & \textbf{0.572} & \textbf{0.224} & \textbf{0.398} \\
    \cmidrule(lr){3-8}
     &  & \multirow{2}{*}{SEER} & \textsc{Uniform}  & 0.260 & 0.553 & 0.210 & 0.382 \\
     &  &  & \textsc{Searched} & \textbf{0.290} & \textbf{0.559} & \textbf{0.226} & \textbf{0.393} \\
    \cmidrule(lr){3-8}
     &  & \multirow{2}{*}{Frequency} & \textsc{Uniform}  & 0.234 & 0.387 & 0.158 & 0.272 \\
     &  &  & \textsc{Searched} & \textbf{0.252} & \textbf{0.405} & \textbf{0.166} & \textbf{0.286} \\
    \cmidrule(lr){3-8}
     &  & \multirow{2}{*}{REAP} & \textsc{Uniform}  & 0.301 & 0.545 & 0.222 & 0.384 \\
     &  &  & \textsc{Searched} & \textbf{0.314} & \textbf{0.571} & \textbf{0.224} & \textbf{0.397} \\
    \midrule[\heavyrulewidth]
    \multirow{21}{*}{\makecell[l]{ERNIE-4.5-\\21B-A3B-PT}} & \multicolumn{3}{l|}{\textbf{Baseline}} & 0.479 & 0.829 & 0.780 & 0.804 \\
    \cmidrule(lr){2-8}
     & \multirow{8}{*}{25\%} & \multirow{2}{*}{EAN} & \textsc{Uniform}  & \textbf{0.377} & 0.815 & 0.748 & 0.781 \\
     &  &  & \textsc{Searched} & 0.333 & \textbf{0.823} & \textbf{0.772} & \textbf{0.797} \\
    \cmidrule(lr){3-8}
     &  & \multirow{2}{*}{SEER} & \textsc{Uniform}  & \textbf{0.301} & \textbf{0.804} & \textbf{0.736} & \textbf{0.770} \\
     &  &  & \textsc{Searched} & 0.291 & \textbf{0.804} & 0.728 & 0.766 \\
    \cmidrule(lr){3-8}
     &  & \multirow{2}{*}{Frequency} & \textsc{Uniform}  & 0.314 & 0.810 & 0.692 & 0.751 \\
     &  &  & \textsc{Searched} & \textbf{0.316} & \textbf{0.815} & \textbf{0.716} & \textbf{0.765} \\
    \cmidrule(lr){3-8}
     &  & \multirow{2}{*}{REAP} & \textsc{Uniform}  & 0.354 & \textbf{0.821} & 0.730 & 0.775 \\
     &  &  & \textsc{Searched} & \textbf{0.376} & 0.814 & \textbf{0.752} & \textbf{0.783} \\
    \cmidrule(lr){2-8}
     & \multirow{8}{*}{50\%} & \multirow{2}{*}{EAN} & \textsc{Uniform}  & 0.156 & \textbf{0.748} & 0.542 & 0.645 \\
     &  &  & \textsc{Searched} & \textbf{0.186} & 0.744 & \textbf{0.558} & \textbf{0.651} \\
    \cmidrule(lr){3-8}
     &  & \multirow{2}{*}{SEER} & \textsc{Uniform}  & 0.130 & 0.555 & 0.418 & 0.487 \\
     &  &  & \textsc{Searched} & \textbf{0.151} & \textbf{0.640} & \textbf{0.508} & \textbf{0.574} \\
    \cmidrule(lr){3-8}
     &  & \multirow{2}{*}{Frequency} & \textsc{Uniform}  & 0.130 & 0.522 & 0.273 & 0.397 \\
     &  &  & \textsc{Searched} & \textbf{0.151} & \textbf{0.631} & \textbf{0.468} & \textbf{0.549} \\
    \cmidrule(lr){3-8}
     &  & \multirow{2}{*}{REAP} & \textsc{Uniform}  & \textbf{0.215} & 0.695 & \textbf{0.598} & 0.646 \\
     &  &  & \textsc{Searched} & 0.205 & \textbf{0.718} & 0.578 & \textbf{0.648} \\
    \midrule[\heavyrulewidth]
    \multirow{21}{*}{\makecell[l]{Qwen3-30B-\\A3B-Instruct-\\2507}} & \multicolumn{3}{l|}{\textbf{Baseline}} & 0.644 & 0.923 & 0.802 & 0.863 \\
    \cmidrule(lr){2-8}
     & \multirow{8}{*}{25\%} & \multirow{2}{*}{EAN} & \textsc{Uniform}  & \textbf{0.517} & \textbf{0.902} & 0.748 & 0.825 \\
     &  &  & \textsc{Searched} & 0.439 & 0.901 & \textbf{0.752} & \textbf{0.827} \\
    \cmidrule(lr){3-8}
     &  & \multirow{2}{*}{SEER} & \textsc{Uniform}  & 0.449 & 0.891 & 0.636 & 0.764 \\
     &  &  & \textsc{Searched} & \textbf{0.482} & \textbf{0.897} & \textbf{0.716} & \textbf{0.806} \\
    \cmidrule(lr){3-8}
     &  & \multirow{2}{*}{Frequency} & \textsc{Uniform}  & 0.446 & 0.891 & 0.658 & 0.774 \\
     &  &  & \textsc{Searched} & \textbf{0.463} & \textbf{0.907} & \textbf{0.734} & \textbf{0.821} \\
    \cmidrule(lr){3-8}
     &  & \multirow{2}{*}{REAP} & \textsc{Uniform}  & 0.565 & 0.910 & 0.784 & 0.847 \\
     &  &  & \textsc{Searched} & \textbf{0.588} & \textbf{0.915} & \textbf{0.806} & \textbf{0.861} \\
    \cmidrule(lr){2-8}
     & \multirow{8}{*}{50\%} & \multirow{2}{*}{EAN} & \textsc{Uniform}  & 0.231 & 0.833 & 0.456 & 0.644 \\
     &  &  & \textsc{Searched} & \textbf{0.243} & \textbf{0.846} & \textbf{0.494} & \textbf{0.670} \\
    \cmidrule(lr){3-8}
     &  & \multirow{2}{*}{SEER} & \textsc{Uniform}  & 0.112 & \textbf{0.605} & 0.144 & \textbf{0.374} \\
     &  &  & \textsc{Searched} & \textbf{0.142} & 0.550 & \textbf{0.220} & \textbf{0.385} \\
    \cmidrule(lr){3-8}
     &  & \multirow{2}{*}{Frequency} & \textsc{Uniform}  & 0.110 & 0.592 & 0.128 & 0.360 \\
     &  &  & \textsc{Searched} & \textbf{0.182} & \textbf{0.697} & \textbf{0.214} & \textbf{0.455} \\
    \cmidrule(lr){3-8}
     &  & \multirow{2}{*}{REAP} & \textsc{Uniform}  & \textbf{0.299} & \textbf{0.872} & \textbf{0.798} & \textbf{0.835} \\
     &  &  & \textsc{Searched} & 0.267 & 0.867 & 0.792 & 0.830 \\
    \bottomrule
  \end{tabular}%
  }
\end{table*}

\begin{table*}[tbp]
  \centering
  \caption{Full results on multiple-choice benchmarks.}
  \label{tab:appendix_mc_full}
  \resizebox{\textwidth}{!}{%
  \begin{tabular}{l l l l | c c c c c c c c c}
    \toprule
    \multicolumn{4}{c}{} & \multicolumn{9}{c}{\textbf{Multiple Choice}} \\
    \textbf{Model} & \textbf{Compression} & \textbf{Method} &  & \textbf{MMLU} & \textbf{ARC-C} & \textbf{ARC-E} & \textbf{HellaSwag} & \textbf{WinoGrande} & \textbf{BoolQ} & \textbf{OpenBookQA} & \textbf{RTE} & \textbf{MC Avg} \\
    \midrule
    \multirow{10}{*}{\makecell[l]{OLMoE-1B-7B-\\0125-Instruct}} & \multicolumn{3}{l|}{\textbf{Baseline}} & 0.534 & 0.490 & 0.758 & 0.808 & 0.684 & 0.766 & 0.470 & 0.711 & 0.653 \\
    \cmidrule(lr){2-13}
     & \multirow{8}{*}{25\%} & \multirow{2}{*}{EAN} & \textsc{Uniform}  & 0.452 & \textbf{0.358} & 0.558 & \textbf{0.630} & \textbf{0.602} & \textbf{0.727} & \textbf{0.336} & \textbf{0.747} & \textbf{0.551} \\
     &  &  & \textsc{Searched} & \textbf{0.454} & 0.349 & \textbf{0.573} & 0.628 & 0.597 & 0.716 & 0.326 & 0.700 & 0.543 \\
    \cmidrule(lr){3-13}
     &  & \multirow{2}{*}{SEER} & \textsc{Uniform}  & \textbf{0.457} & \textbf{0.352} & 0.566 & 0.620 & 0.599 & \textbf{0.723} & \textbf{0.340} & \textbf{0.708} & \textbf{0.545} \\
     &  &  & \textsc{Searched} & 0.447 & 0.338 & \textbf{0.573} & \textbf{0.621} & \textbf{0.601} & 0.715 & 0.330 & 0.690 & 0.539 \\
    \cmidrule(lr){3-13}
     &  & \multirow{2}{*}{Frequency} & \textsc{Uniform}  & 0.450 & \textbf{0.351} & \textbf{0.572} & \textbf{0.621} & \textbf{0.605} & \textbf{0.727} & \textbf{0.334} & \textbf{0.718} & \textbf{0.547} \\
     &  &  & \textsc{Searched} & \textbf{0.455} & 0.347 & 0.570 & 0.612 & 0.595 & 0.719 & 0.318 & 0.693 & 0.539 \\
    \cmidrule(lr){3-13}
     &  & \multirow{2}{*}{REAP} & \textsc{Uniform}  & \textbf{0.475} & 0.418 & 0.625 & 0.680 & 0.637 & \textbf{0.729} & 0.362 & \textbf{0.704} & 0.579 \\
     &  &  & \textsc{Searched} & 0.474 & \textbf{0.427} & \textbf{0.641} & \textbf{0.685} & \textbf{0.639} & 0.712 & \textbf{0.386} & 0.682 & \textbf{0.581} \\
    \midrule[\heavyrulewidth]
    \multirow{10}{*}{\makecell[l]{DeepSeek-V2-Lite-\\Chat}} & \multicolumn{3}{l|}{\textbf{Baseline}} & 0.567 & 0.541 & 0.785 & 0.808 & 0.712 & 0.829 & 0.456 & 0.726 & 0.678 \\
    \cmidrule(lr){2-13}
     & \multirow{8}{*}{25\%} & \multirow{2}{*}{EAN} & \textsc{Uniform}  & 0.435 & 0.404 & 0.623 & 0.666 & \textbf{0.679} & 0.771 & 0.324 & 0.668 & 0.571 \\
     &  &  & \textsc{Searched} & \textbf{0.468} & \textbf{0.410} & \textbf{0.649} & \textbf{0.685} & 0.670 & \textbf{0.787} & \textbf{0.336} & \textbf{0.700} & \textbf{0.588} \\
    \cmidrule(lr){3-13}
     &  & \multirow{2}{*}{SEER} & \textsc{Uniform}  & 0.431 & 0.428 & \textbf{0.660} & 0.671 & 0.660 & \textbf{0.770} & 0.342 & \textbf{0.722} & 0.586 \\
     &  &  & \textsc{Searched} & \textbf{0.457} & \textbf{0.433} & \textbf{0.660} & \textbf{0.693} & \textbf{0.668} & \textbf{0.770} & \textbf{0.362} & 0.708 & \textbf{0.594} \\
    \cmidrule(lr){3-13}
     &  & \multirow{2}{*}{Frequency} & \textsc{Uniform}  & 0.427 & 0.389 & 0.611 & 0.647 & 0.645 & 0.753 & 0.332 & 0.675 & 0.560 \\
     &  &  & \textsc{Searched} & \textbf{0.484} & \textbf{0.397} & \textbf{0.647} & \textbf{0.679} & \textbf{0.657} & \textbf{0.770} & \textbf{0.352} & \textbf{0.718} & \textbf{0.588} \\
    \cmidrule(lr){3-13}
     &  & \multirow{2}{*}{REAP} & \textsc{Uniform}  & 0.465 & 0.457 & 0.662 & 0.738 & \textbf{0.686} & 0.770 & 0.378 & 0.679 & 0.604 \\
     &  &  & \textsc{Searched} & \textbf{0.502} & \textbf{0.490} & \textbf{0.726} & \textbf{0.766} & 0.676 & \textbf{0.795} & \textbf{0.414} & \textbf{0.740} & \textbf{0.639} \\
    \midrule[\heavyrulewidth]
    \multirow{21}{*}{\makecell[l]{ERNIE-4.5-\\21B-A3B-PT}} & \multicolumn{3}{l|}{\textbf{Baseline}} & 0.739 & 0.564 & 0.782 & 0.814 & 0.717 & 0.872 & 0.462 & 0.816 & 0.721 \\
    \cmidrule(lr){2-13}
     & \multirow{8}{*}{25\%} & \multirow{2}{*}{EAN} & \textsc{Uniform}  & 0.625 & 0.496 & 0.710 & 0.719 & 0.705 & 0.862 & \textbf{0.408} & \textbf{0.827} & 0.669 \\
     &  &  & \textsc{Searched} & \textbf{0.646} & \textbf{0.509} & \textbf{0.725} & \textbf{0.731} & \textbf{0.708} & \textbf{0.864} & 0.404 & 0.809 & \textbf{0.675} \\
    \cmidrule(lr){3-13}
     &  & \multirow{2}{*}{SEER} & \textsc{Uniform}  & 0.628 & \textbf{0.466} & 0.687 & 0.655 & 0.655 & \textbf{0.856} & 0.342 & \textbf{0.780} & 0.634 \\
     &  &  & \textsc{Searched} & \textbf{0.638} & \textbf{0.466} & \textbf{0.693} & \textbf{0.663} & \textbf{0.665} & 0.843 & \textbf{0.368} & 0.773 & \textbf{0.638} \\
    \cmidrule(lr){3-13}
     &  & \multirow{2}{*}{Frequency} & \textsc{Uniform}  & 0.608 & 0.460 & 0.682 & 0.669 & \textbf{0.671} & 0.846 & 0.364 & 0.791 & 0.636 \\
     &  &  & \textsc{Searched} & \textbf{0.632} & \textbf{0.480} & \textbf{0.700} & \textbf{0.676} & 0.662 & \textbf{0.850} & \textbf{0.382} & \textbf{0.794} & \textbf{0.647} \\
    \cmidrule(lr){3-13}
     &  & \multirow{2}{*}{REAP} & \textsc{Uniform}  & \textbf{0.643} & \textbf{0.521} & 0.756 & 0.718 & 0.692 & \textbf{0.855} & 0.404 & 0.747 & 0.667 \\
     &  &  & \textsc{Searched} & 0.638 & 0.510 & \textbf{0.764} & \textbf{0.719} & \textbf{0.698} & 0.852 & \textbf{0.408} & \textbf{0.787} & \textbf{0.672} \\
    \cmidrule(lr){2-13}
     & \multirow{8}{*}{50\%} & \multirow{2}{*}{EAN} & \textsc{Uniform}  & \textbf{0.510} & 0.417 & \textbf{0.623} & \textbf{0.575} & 0.631 & 0.836 & \textbf{0.328} & \textbf{0.762} & \textbf{0.585} \\
     &  &  & \textsc{Searched} & 0.508 & \textbf{0.424} & 0.622 & 0.568 & \textbf{0.642} & \textbf{0.839} & 0.318 & 0.736 & 0.582 \\
    \cmidrule(lr){3-13}
     &  & \multirow{2}{*}{SEER} & \textsc{Uniform}  & 0.494 & \textbf{0.396} & 0.579 & \textbf{0.505} & \textbf{0.606} & \textbf{0.811} & \textbf{0.300} & \textbf{0.718} & \textbf{0.551} \\
     &  &  & \textsc{Searched} & \textbf{0.503} & 0.394 & \textbf{0.595} & 0.484 & 0.591 & 0.800 & 0.298 & 0.708 & 0.547 \\
    \cmidrule(lr){3-13}
     &  & \multirow{2}{*}{Frequency} & \textsc{Uniform}  & \textbf{0.519} & \textbf{0.406} & \textbf{0.592} & \textbf{0.521} & \textbf{0.597} & \textbf{0.829} & 0.292 & \textbf{0.765} & \textbf{0.565} \\
     &  &  & \textsc{Searched} & 0.508 & 0.397 & 0.586 & 0.497 & 0.575 & 0.821 & \textbf{0.316} & 0.755 & 0.557 \\
    \cmidrule(lr){3-13}
     &  & \multirow{2}{*}{REAP} & \textsc{Uniform}  & 0.502 & 0.407 & 0.622 & \textbf{0.556} & 0.625 & \textbf{0.814} & \textbf{0.344} & \textbf{0.733} & \textbf{0.575} \\
     &  &  & \textsc{Searched} & \textbf{0.508} & \textbf{0.410} & \textbf{0.637} & 0.554 & \textbf{0.627} & 0.804 & 0.334 & 0.726 & \textbf{0.575} \\
    \midrule[\heavyrulewidth]
    \multirow{21}{*}{\makecell[l]{Qwen3-30B-\\A3B-Instruct-\\2507}} & \multicolumn{3}{l|}{\textbf{Baseline}} & 0.802 & 0.625 & 0.838 & 0.797 & 0.736 & 0.887 & 0.446 & 0.769 & 0.737 \\
    \cmidrule(lr){2-13}
     & \multirow{8}{*}{25\%} & \multirow{2}{*}{EAN} & \textsc{Uniform}  & \textbf{0.635} & 0.451 & 0.636 & \textbf{0.655} & 0.676 & \textbf{0.862} & 0.338 & \textbf{0.769} & 0.628 \\
     &  &  & \textsc{Searched} & 0.598 & \textbf{0.491} & \textbf{0.674} & 0.645 & \textbf{0.692} & 0.857 & \textbf{0.354} & 0.758 & \textbf{0.634} \\
    \cmidrule(lr){3-13}
     &  & \multirow{2}{*}{SEER} & \textsc{Uniform}  & \textbf{0.533} & 0.372 & 0.457 & 0.609 & 0.597 & 0.849 & 0.306 & \textbf{0.765} & 0.561 \\
     &  &  & \textsc{Searched} & 0.495 & \textbf{0.400} & \textbf{0.509} & \textbf{0.637} & \textbf{0.629} & \textbf{0.859} & \textbf{0.344} & 0.744 & \textbf{0.577} \\
    \cmidrule(lr){3-13}
     &  & \multirow{2}{*}{Frequency} & \textsc{Uniform}  & \textbf{0.527} & 0.370 & 0.457 & 0.611 & 0.582 & 0.846 & 0.312 & \textbf{0.762} & 0.558 \\
     &  &  & \textsc{Searched} & 0.494 & \textbf{0.404} & \textbf{0.500} & \textbf{0.630} & \textbf{0.616} & \textbf{0.852} & \textbf{0.340} & 0.740 & \textbf{0.572} \\
    \cmidrule(lr){3-13}
     &  & \multirow{2}{*}{REAP} & \textsc{Uniform}  & \textbf{0.734} & \textbf{0.584} & \textbf{0.803} & \textbf{0.737} & \textbf{0.733} & \textbf{0.883} & 0.404 & \textbf{0.773} & \textbf{0.706} \\
     &  &  & \textsc{Searched} & 0.733 & 0.583 & 0.794 & 0.733 & 0.732 & \textbf{0.883} & \textbf{0.416} & 0.754 & 0.704 \\
    \cmidrule(lr){2-13}
     & \multirow{8}{*}{50\%} & \multirow{2}{*}{EAN} & \textsc{Uniform}  & 0.464 & \textbf{0.341} & \textbf{0.475} & 0.483 & \textbf{0.594} & 0.791 & 0.298 & 0.679 & 0.516 \\
     &  &  & \textsc{Searched} & \textbf{0.465} & 0.322 & 0.473 & \textbf{0.489} & 0.590 & \textbf{0.803} & \textbf{0.310} & \textbf{0.690} & \textbf{0.518} \\
    \cmidrule(lr){3-13}
     &  & \multirow{2}{*}{SEER} & \textsc{Uniform}  & \textbf{0.421} & \textbf{0.284} & 0.352 & \textbf{0.448} & 0.524 & \textbf{0.702} & 0.282 & \textbf{0.625} & \textbf{0.455} \\
     &  &  & \textsc{Searched} & 0.412 & 0.274 & \textbf{0.357} & 0.438 & \textbf{0.538} & 0.695 & \textbf{0.284} & 0.567 & 0.446 \\
    \cmidrule(lr){3-13}
     &  & \multirow{2}{*}{Frequency} & \textsc{Uniform}  & 0.421 & 0.294 & 0.346 & 0.447 & 0.515 & 0.695 & 0.284 & 0.596 & 0.450 \\
     &  &  & \textsc{Searched} & \textbf{0.431} & \textbf{0.309} & \textbf{0.498} & \textbf{0.500} & \textbf{0.593} & \textbf{0.754} & \textbf{0.306} & \textbf{0.686} & \textbf{0.510} \\
    \cmidrule(lr){3-13}
     &  & \multirow{2}{*}{REAP} & \textsc{Uniform}  & \textbf{0.591} & \textbf{0.459} & \textbf{0.652} & \textbf{0.543} & \textbf{0.668} & \textbf{0.813} & 0.328 & 0.718 & \textbf{0.596} \\
     &  &  & \textsc{Searched} & 0.576 & 0.437 & 0.638 & 0.518 & 0.629 & 0.805 & \textbf{0.332} & \textbf{0.747} & 0.585 \\
    \bottomrule
  \end{tabular}%
  }
\end{table*}

\Cref{tab:appendix_code_full,tab:appendix_wild_math_full,tab:appendix_mc_full} provide the benchmark-level breakdown behind the averaged results reported in \cref{tab:main_table_olmoe_ernie,tab:main_table_deepseek_qwen3}. Specifically, they expand Code Avg into HumanEval, HumanEval$+$, MBPP, MBPP$+$, and LiveCodeBench; Math Avg into GSM8K and MATH-500 (with WildBench reported alongside them); and MC Avg into its eight constituent multiple-choice benchmarks.

\end{document}